\documentclass[runningheads]{llncs}
\usepackage{graphicx}

\usepackage{tikz}
\usepackage{comment}
\usepackage{amsmath,amssymb} %
\usepackage{color}
\usepackage{dsfont}

\usepackage{xcolor}
\usepackage{graphicx}
\usepackage{float}
\usepackage{multicol}
\usepackage{multirow}
\usepackage{caption}
\usepackage{subcaption}
\usepackage{booktabs}
\usepackage{xparse}
\usepackage{xspace}
\usepackage{paralist}
\usepackage{wrapfig}
\usepackage{caption}
\usepackage{subcaption}
\usepackage{layouts}
\usepackage{enumitem}
\usepackage{tabularx}
\usepackage[pagebackref,breaklinks,colorlinks]{hyperref}

\usepackage[accsupp]{axessibility}  %

\definecolor{myOrange}{rgb}{1,0.5,0}
\definecolor{myBlue}{rgb}{0.1,0.2,0.8}
\definecolor{myGreen}{rgb}{0.1,0.8,0.1}

\newcommand{\myparagraph}[1]{\vspace{0.75ex}\par\noindent \textbf{#1}}

\newcommand{\eg}{\textit{e}.\textit{g}.\ }

\newcommand{\notation}[1]{\ensuremath{#1}\xspace}

\newcommand{\GANCriterion}{\notation{V}}
\newcommand{\Generator}{\notation{G}}
\newcommand{\GeneratorFixedScale}{\notation{G_{\text{fixed}}}}
\newcommand{\Fourier}{\notation{F}}
\newcommand{\AuxNetwork}{\notation{M}}

\newcommand{\Mask}{\notation{m}}
\newcommand{\Distance}{\notation{d}}
\newcommand{\Warp}{\notation{w_{\PatchLocation,\FullImageResolution}}}

\newcommand{\ImageSample}{\notation{x}}
\newcommand{\Dataset}{\notation{\mathcal{D}}}
\newcommand{\FixedDataset}{\notation{\mathcal{D}_{\text{fixed}}}}

\newcommand{\Discriminator}{\notation{D}}

\newcommand{\PatchResolution}{\notation{p}}
\newcommand{\NativeResolution}{\notation{s_\text{im}}}
\newcommand{\FullImageResolution}{\notation{s}}
\newcommand{\NormalizedFullImageResolution}{\notation{\bar{\FullImageResolution}}}
\newcommand{\MaxNormImageResolution}{\notation{s_\text{max}}}
\newcommand{\ContinuousDomain}{\notation{[0,1]\times[0,1]}}
\newcommand{\LatentCode}{\notation{z}}
\newcommand{\StyleCode}{\notation{w}}
\newcommand{\PatchLocation}{\notation{\mathbf{v}}}
\newcommand{\PatchLocationX}{\notation{\PatchLocation_x}}
\newcommand{\PatchLocationY}{\notation{\PatchLocation_y}}
\newcommand{\Coordinates}{\notation{c_{\PatchLocation,\FullImageResolution}}}
\newcommand{\NormalizedCoordinates}{\notation{c}}
\newcommand{\AvgResolution}{\notation{\mathbb{E}[s]}}
\newcommand{\TransformPatch}{\notation{T_{\mathrm{patch}}}}

\begin{document}
\pagestyle{headings}
\mainmatter
\def\ECCVSubNumber{3693}  %

\title{Any-resolution Training for \\ High-resolution Image Synthesis}

\titlerunning{Anyres-GAN}
\author{Lucy Chai\inst{1} \and Michaël Gharbi\inst{2} \and Eli Shechtman\inst{2} \\ Phillip Isola\inst{1} \and Richard Zhang\inst{2}}
\authorrunning{L. Chai et al.}
\institute{\textsuperscript{1}MIT \enspace \textsuperscript{2}Adobe Research}
\maketitle

\begin{abstract}
    Generative models operate at fixed resolution, even though
    natural images come in a variety of sizes.
    As high-resolution details are downsampled away and low-resolution images
    are discarded altogether, precious supervision is lost.
    We argue that every pixel matters and create datasets with variable-size
    images, collected at their native resolutions.
    To take advantage of varied-size data, 
    we introduce \textit{continuous-scale} training, a process that samples
    \textit{patches} at random scales to train a new generator with variable
    output resolutions.
    First, conditioning the generator on a target scale allows us to generate
    higher resolution images than previously possible, without adding layers to the
    model.
    Second, by conditioning on continuous coordinates, we can sample patches that
    still obey a consistent global layout, which also allows for scalable training at higher resolutions.
    Controlled FFHQ experiments show that our method can take advantage of
    multi-resolution training data better than discrete multi-scale
    approaches, achieving better FID scores and cleaner 
    high-frequency details.
    We also train on other natural image domains including churches,
    mountains, and birds, and demonstrate arbitrary scale synthesis with
    both coherent global layouts and realistic local details,
    going beyond 2K resolution in our experiments. Our project page is available at: \url{https://chail.github.io/anyres-gan/}.

\end{abstract}

\begin{figure}[t]
\centering
\includegraphics[width=\textwidth]{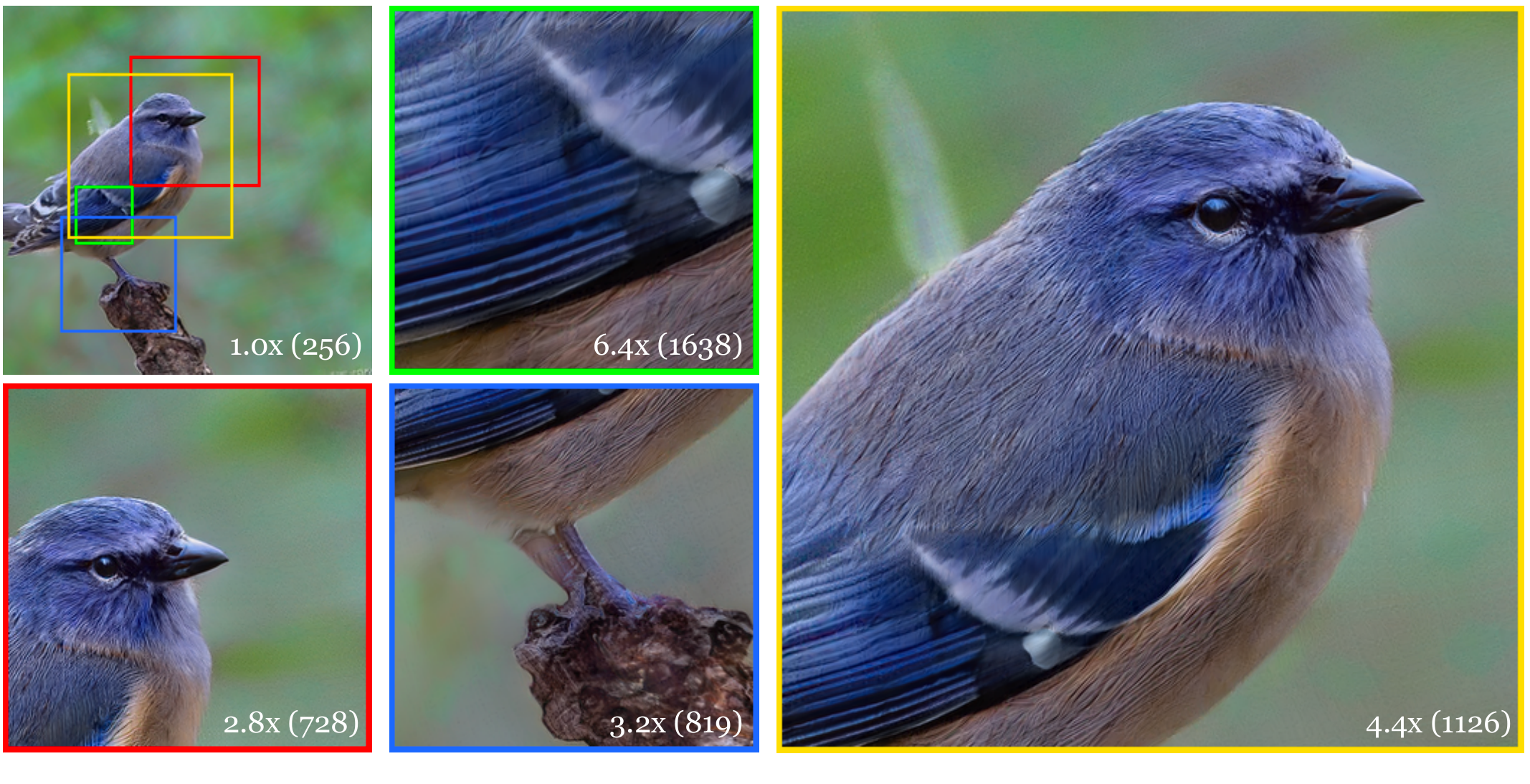}
\vspace{-7mm}
\caption{\small
Trained on a dataset of varied-size images, our unconditional generator learns to
synthesize patches at continuous scales to match the distribution of real
patches.
Here, we render crops of the image at different resolutions, indicating the target resolution for each.
We indicate the region of each crop in the top-left panel, which is the image directly sampled without scale input.
\vspace{-5mm}
}
\label{fig:teaser}
\end{figure}

\section{Introduction}

The first step of typical generative modeling pipelines is to build a dataset with a fixed, target resolution. Images above the target resolution are downsampled, removing high-frequency details,
and data of insufficient resolution is
omitted, discarding structural information about low frequencies.
Our insight is that this process wastes potentially learnable information.
We propose to embrace the natural diversity of image sizes, processing them at their \emph{native} resolution.

Relaxing the fixed-dataset assumption offers new, previously unexplored opportunities. One can potentially simultaneously learn global structure -- for which large sets of readily-available low-resolution images suffice -- and fine-scale details -- where even a handful of high-resolution images may be adequate, especially given their internal recurrence~\cite{shocher2018zero}.
This enables generating images at higher resolutions than previously possible, by adding in higher-resolution images to currently available fixed-size datasets.

This problem setting offers unique challenges, both in regards to modeling and scaling. First, one must generate images across multiple scales in order to compare with the target distribution. Na\"{i}vely downsampling the full-resolution image is suboptimal, as it is common to even have images of $8\times$ difference in scale in the dataset.
Secondly, generating and processing high-resolution images offers scaling challenges. Modern training architectures at 1024 resolution already push the current hardware to the limits in memory and training time, and are unable to fully make use of images above that resolution. 

To bypass these issues, we design a generator to synthesize image crops at arbitrary scales, hence, performing \textit{any-resolution} training.
We modify the state-of-the-art StyleGAN3~\cite{karras2021alias} architecture to take a grid of continuous coordinates, defined on a bounded domain, as well as a target scale, inspired by recent work in coordinate-based conditioning~\cite{mildenhall2020nerf,barron2021mip,lin2021infinitygan,chen2021learning}.
By keeping the latent code constant and varying the crop coordinates and scale, our generator can output patches of the same image~\cite{lin2019coco,shaham2019singan}, but at various scales. This allows us to (1) efficiently generate at arbitrary scale, so that a discriminator critic can compare generations to a multi-resolution dataset, and (2) decouple high-resolution synthesis from increasing architecture size and prohibitive memory requirements.

We first experiment on FFHQ images~\cite{karras2017progressive} as a controlled setting, showing efficient data usage by comparing our training on image subsets to using the entire full resolution dataset. We find minimal degradation (FIDs varying by 0.3), even at highly skewed distributions -- $98\%$ of the dataset at $4\times$ lower resolution, and just $2\%$ at a higher, mixed resolutions. %
Practically, this means we can leverage large-scale ($>$100k) lower-resolution datasets, such as LSUN Churches~\cite{yu15lsun}, Flickr Mountains~\cite{park2020swapping}, and Flickr Birds collected by us, and add a relatively small amount of high-resolution images ($\sim6000$), for continuous resolution synthesis beyond the 1024 resolution limit of current generators. 
To summarize, we:
\begin{itemize}[topsep=0pt,leftmargin=*]
    \item propose to train on mixed-resolution datasets from images in-the-wild.
    \item modify the generator to be amenable to such data, sampling patches at arbitrary scales during our \textit{any-resolution} training procedure.
    \item demonstrate successful generations beyond $1024\times 1024$, with fine details and coherent global structure, without a larger and more expensive generator.
    \item introduce a variant of the FID metric that captures image statistics at multiple scales, thus accounting for the details of high resolutions.
\end{itemize}

\section{Related Works}

\myparagraph{Unconditional image synthesis.}
Recent generative models including
GANs \cite{goodfellow2014explaining,karras2019style,karras2021alias,brock2018large},
Variational Autoencoders~\cite{kingma2013auto}, diffusion
models~\cite{nichol2021improved,ho2020denoising,song2020denoising,song2020improved,dhariwal2021diffusion},
and autoregressive
models~\cite{larochelle2011neural,van2016pixel,van2016conditional} such as
transformers~\cite{vaswani2017attention,chen2020generative,esser2021taming}
are rapidly improving in quality. Of these, we focus on GANs, which offer
state-of-the-art performance along with efficient inference and effective editing properties.
A key innovation in GANs has been multi-resolution supervision during training. Works such as LapGAN~\cite{denton2015deep}, the Progressive/StyleGAN family~\cite{karras2017progressive,karras2019style,karras2019analyzing,karras2021alias}, MSG-GAN~\cite{karnewar2020msg}, and AnyCost-GAN~\cite{lin2021anycost} have demonstrated stable training by
growing the generator with additional layers that increase resolution by factors of two.
Such a strategy even works for single-image GANs~\cite{shaham2019singan,shocher2019ingan}, based on the observation that images share statistics across scales.
While several works~\cite{karras2020training,zhao2020image,zhao2020differentiable} show data augmentations, such as small jitters in scale, can help stabilize training, they are processing the same, underlying fixed-resolution dataset. We draw upon the insights in these works for stable training, and seek to unlock training on an any-resolution dataset.
Importantly, our generator does not use additional layers and can synthesize images at continuous scales, not only powers of two.

\myparagraph{Coordinate-based functions.}
Coordinate-based encodings enable spatial conditioning and
provide an inductive bias towards natural images~\cite{tancik2020fourier}.
Recent methods use point-based neural mappings to transform 2D or 3D coordinates
to a color value for the purposes of unconditional
generation~\cite{lin2019coco,choi2021toward,karras2021alias,skorokhodov2021adversarial,lin2021infinitygan,anokhin2021image},
conditional generation~\cite{shaham2021spatially}, 3D view
synthesis~\cite{mildenhall2020nerf,schwarz2020graf,chan2021pi}, or fitting arbitrary
signals~\cite{chen2021learning,mehta2021modulated}.
By oversampling the coordinate grid, one can generate a larger image at
inference time. However, because these models keep the same fixed-scale dataset
assumption during training, the outputs struggle to offer additional
high-frequency details without a high-frequency training signal. We draw upon the
innovations in coordinate-based functions to sample patches at different scales
and locations, enabling us to efficiently train on multiple scales.
MS-PIE~\cite{xu2021positional} and MS-PE~\cite{choi2021toward} add positional encodings for multi-scale synthesis, but retain a global image discriminator at smaller resolution. Concurrently, ScaleParty~\cite{ntavelis2022arbitrary} also samples patches, but their goal is to generate at multiple scales with cross-scale consistency while we focus on training with arbitrary size real images.

\myparagraph{Extrapolation.}
One method of generating ``infinite'' resolution is extrapolating an image.
Early texture synthesis works~\cite{efros1999texture,efros2001image,wexler2007space} focus on stationary textures. Recent advances~\cite{zhou2018non} explore non-stationary textures, with large-scale structures and inhomogeneous patterns. Similar approaches operate by outpainting images, extending images beyond their boundaries in a conditional setting~\cite{teterwak2019boundless,yang2019very,ma2021boosting,zhou2018non}. Recent generative models synthesize large scenes~\cite{lin2021infinitygan,cheng2021out}, typically casting synthesis as an outpainting problem~\cite{yang2019very,ma2021boosting}. These methods are most effective for signals with a strong stationary component, such as landscapes, although extrapolation of structured scenes can be achieved in some domains~\cite{wang2019wide}. %
Unlike textures, the images we wish to synthesize typically have a strong global structure. In a sense, we seek to extrapolate by ``zooming in'' or out, rather than ``panning'' beyond an image's boundaries.

\myparagraph{Super-resolution.}
An alternative approach to generating high-resolution imagery would be to start with an off-the-shelf generative model and feed its outputs to a super-resolution method~\cite{hu2019meta,chen2021learning,wang2018esrgan,wang2021real},
possibly exploiting the self-similarity properties of
images~\cite{shechtman2007matching,glasner2009super,huang2015single,shocher2018zero}.
Applying super-resolution models is challenging, in part because of the specific
blur kernels super-resolution models are trained on~\cite{zhang2020aim}.
Furthermore, though generations continue to improve, there remains a persistent domain shift between synthesized and real images~\cite{wang2020cnn,chai2020makes}.
Finally, super-resolution is a local, conditional problem where the global structure
is dictated by the low-resolution input, and optionally an additional high-resolution reference image~\cite{zheng2018crossnet,yang2020learning,lu2021masa,jiang2021robust,xia2022coarse}.
We synthesize plausible images unconditionally, leveraging a \textit{set} of high-resolution images to produce both realistic global structure and coherent fine details.

\begin{wrapfigure}{r}{0.35\textwidth}
    \vspace{-15mm}
  \begin{center}
    \includegraphics[width=.35\textwidth]{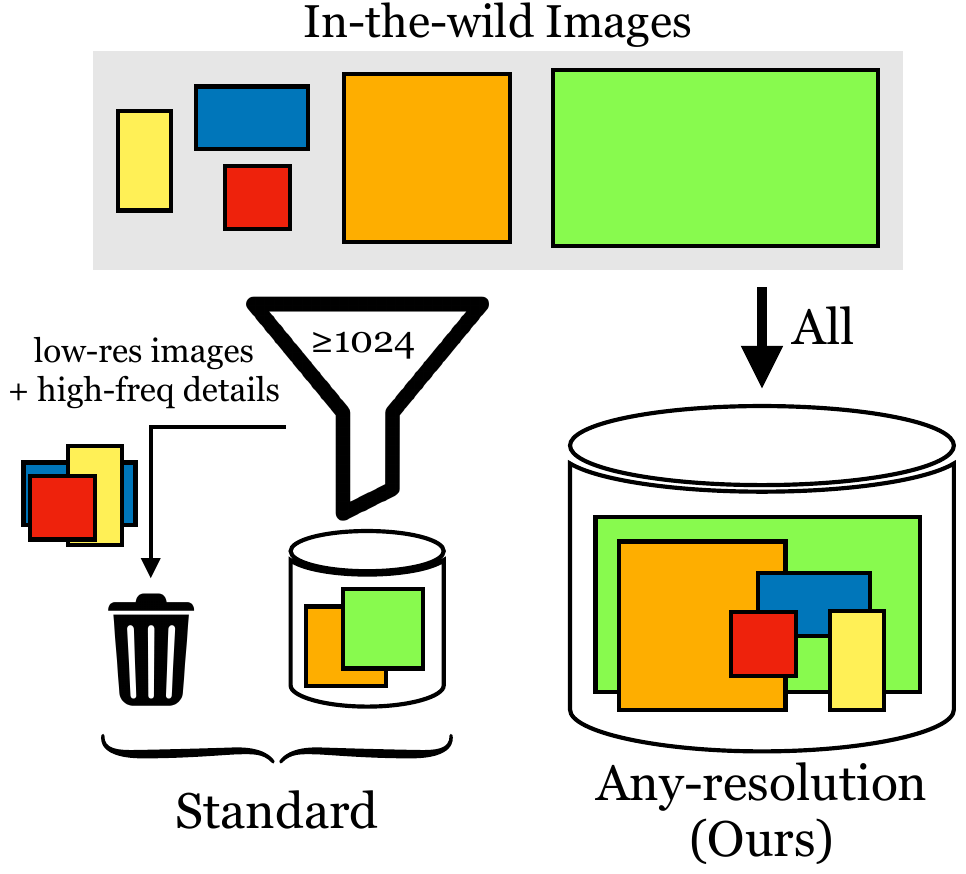}
  \end{center}
  \vspace{-8mm}
  \caption{\small Data comes at a variety of scales. Traditional dataset
  construction filters out low-resolution images and downsamples
  high-resolution images to a fixed, training resolution. We aim to keep
  images at their original resolution.}
  \vspace{-10mm}
  \label{fig:dataset_construction}
\end{wrapfigure}

\section{Methods}\label{sec:method}

In standard GAN training, all training images share a common fixed resolution,
which matches the generator's output size.
We seek to exploit the variety of image resolutions available in the
wild, learning from pixels that are usually discarded, to enable
high- and continuously-variable resolution synthesis.
We achieve this by switching from the common fixed-resolution thinking, to a
novel `any-resolution' approach, where the original size of each training image
is preserved (Fig~\ref{fig:dataset_construction}).
We introduce a new class of GAN generators that learn from this multi-resolution
signal to synthesize images at any resolution
(\S~\ref{sec:multiresolution_gan}), and show how to train them by sampling
patches at multiple scales to jointly supervise the global-structure and fine
image details (\S~\ref{sec:training}).

\begin{figure*}[t]
    \centering
    \includegraphics[width=1.\linewidth]{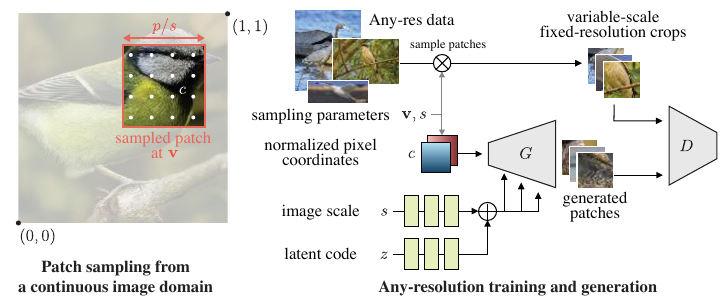}
    \vspace{-5mm}
    \caption{\label{fig:schematic}
    \small Overview. 
    (Left) We parameterize images (real or synthetic) as continuous functions
    over a normalized domain and extract random patches at various
    scales \FullImageResolution, but constant resolution \PatchResolution.
    (Right) To train, we sample crops at random scales and offsets \PatchLocation from
    full-size real images.
    The same crops are sampled from the generator, by passing
    it a grid of the desired coordinates \Coordinates,
    and injecting the image scale \FullImageResolution through modulation,
    in addition to a global latent code \LatentCode.
    }
    \vspace{-5mm}
\end{figure*}

\subsection{Multi-resolution GAN}\label{sec:multiresolution_gan}
We design our approach to leverage state-of-the-art GANs.
We keep the architecture of the discriminator
unchanged.
Since the discriminator operates at fixed resolution, we modify the
generator to synthesize images at any resolution and receive
the discriminator's fixed-resolution supervision.
Our implementation builds on the StyleGAN3 framework~\cite{karras2021alias}, which is conditioned on a fixed coordinate grid.
We modify this grid for any-resolution and patch-based synthesis.

\myparagraph{Continuous-resolution generator.}
We treat each image as a continuous function defined on a bounded normalized
coordinate domain \ContinuousDomain.
The generator \Generator always generates patches at a fixed pixel resolution
$\PatchResolution\times\PatchResolution$, but each patch implicitly corresponds to a square
sub-region, centered at $\PatchLocation\in[0,1]^2$, of the larger image.
Denoting the resolution of the larger image as $\FullImageResolution\times\FullImageResolution $, we have that the patch size is $\PatchResolution/\FullImageResolution$ in normalized coordinates (see Figure~\ref{fig:schematic}, left).
During training, we sample patches from images at multiple scales
\FullImageResolution, either from the generator or from the multi-resolution
dataset, before passing them to the fixed-resolution discriminator \Discriminator.
Formally, our generator takes three inputs: a regular grid of normalized continuous pixel
coordinates $\Coordinates\in\mathds{R}^{\PatchResolution\times
\PatchResolution\times 2}$, the
resolution $\FullImageResolution \in \mathds{N}$ of the (implicit) larger image the patch is extracted from, and \LatentCode, the latent code representing this larger image.
It synthesizes the patch's pixel values at the sampled coordinates as:
\begin{equation}
\Generator(\LatentCode,\Coordinates,\FullImageResolution) = \Generator(\Fourier(\Coordinates); \AuxNetwork(\LatentCode, \FullImageResolution)),
\end{equation}
where \Fourier is a Fourier embedding of the continuous
coordinates~\cite{karras2021alias}, and \AuxNetwork is an auxiliary function
that maps the latent code and sampling resolution into a set of modulation
parameters for the StyleGAN3 generator (see \S~\ref{sec:details} for details).
Our method therefore modifies two components from StyleGAN3.
First, we replace the fixed coordinate grid with \emph{patch-dependent}
coordinates to train on variable-resolution images; these coordinates are adjusted to account for upsampling in StyleGAN3.
Second, we append an auxiliary branch \AuxNetwork to inject scale information
throughout the generator.  

At test time, we can generate images at arbitrarily high resolutions by sampling
the full continuous domain \ContinuousDomain at the desired sampling rate.
Theoretically, the maximum resolution is infinite, but in practice the amount of detail that the model can generate is determined by generator resolution \PatchResolution and the resolutions of the training images.

\subsection{Two-phase training}\label{sec:training}

We train our generator in two phases.
In the first, we want the generator to learn to generate globally-coherent
images.
For this, we disable the patch sampling mechanism and pretrain the generator at a
fixed scale, corresponding to the full continuous image domain.
That is, we fix $\FullImageResolution=\PatchResolution$, and
$\PatchLocation=(0.5,0.5)$, 
which is equivalent to standard fixed-resolution GAN training.
Both the coordinate tensor \Coordinates and the scale conditioning
variable \FullImageResolution are constant in this phase, so
we simply refer to the image generated as
$\GeneratorFixedScale(\LatentCode)$ and follow the training procedure of StyleGAN3. %
In the second phase, we enable patch sampling for both the real
and synthetic images and continue training the generator using variable-scale
patches, so it learns to synthesize fine details at any resolution.
We found that using a copy of the pretrained fixed-scale generator
\GeneratorFixedScale as a teacher model helps stabilize training in this phase.

\myparagraph{Global fixed-resolution pretraining.}
During the pretraing phase, we effectively resample all the training images to a
fixed resolution $\PatchResolution\times\PatchResolution$, as in standard GAN training.
Let $\ImageSample\sim\FixedDataset$ denote an image sampled from this fixed size dataset.
We optimize a standard GAN objective with non-saturating logistic loss and $R_1$
regularization on the discriminator: 
\begin{equation}
\begin{split}
    &\GANCriterion(\Discriminator, \Generator(\LatentCode), \ImageSample) = \Discriminator(\ImageSample)-\Discriminator(\Generator(\LatentCode)), \hspace{4mm} R_1(\Discriminator, \ImageSample) = || \nabla \Discriminator(x) || ^2, \\
    & \GeneratorFixedScale = \arg \min_\Generator \max_\Discriminator
     \hspace{1mm} \mathds{E}_{\LatentCode, \ImageSample\sim \FixedDataset}
      \hspace{1mm} \GANCriterion(\Discriminator, \Generator(\LatentCode), \ImageSample) - \frac{\lambda_{R_1}}{2} R_1(\Discriminator, \ImageSample).
\end{split}
\end{equation}
We follow the recommended values for $\lambda_{R_1}$, depending on the 
generator resolution \PatchResolution~\cite{karras2021alias}.

\myparagraph{Mixed-resolution patch-based training.}
In the second phase, we enable multi-resolution sampling, alternating between
extracting random crops from our any-resolution dataset and generating them with
our continuous generator.

For synthetic patches, we sample a patch location \PatchLocation
uniformly in the continuous domain \ContinuousDomain; and
an arbitrary image resolution $\FullImageResolution \geq \PatchResolution$,
corresponding to the implicit full image around the square patch.
From those, we derive the sampling coordinate grid \Coordinates, and synthesize
the patch image $\Generator(\LatentCode, \Coordinates, \FullImageResolution)$, as
described earlier.

For `real' patches, we sample an image from our dataset.
Because this image can have any resolution $\NativeResolution \geq \PatchResolution$, we crop it to a random square matching its smallest dimension, then
Lanczos downsample this square to a random resolution $\FullImageResolution \times \FullImageResolution$ with $\NativeResolution \geq \FullImageResolution \geq \PatchResolution$. 
Finally, we extract a random $\PatchResolution \times \PatchResolution$ crop from
the downsampled image, recording its center \PatchLocation.
To preserve the generator's global coherence and continuous generation ability, we sample at global scale $\FullImageResolution=\PatchResolution$ and $\PatchLocation=(0.5,0.5)$ (similar to the pretraining step) with
probability $50\%$.
We found that image quality at global resolution  $\FullImageResolution=\PatchResolution$ degrades otherwise, 
and we refer to these generated full images of size $\PatchResolution \times \PatchResolution$ as ``base images.'' 
Our any-resolution GAN optimizes the following objective during this phase:
\begin{equation}
\begin{split}
\Generator^{*} = \arg \min_\Generator \min_\Discriminator \hspace{1mm}
\mathds{E}_{\LatentCode, {\{\ImageSample, \FullImageResolution, \PatchLocation\}\sim \mathcal{\Discriminator}}} \hspace{1mm}
 & \GANCriterion(\Discriminator, \Generator(\LatentCode,\Coordinates,\FullImageResolution), \ImageSample) \\ \vspace{-1mm}
+ \lambda_\text{teacher} & \mathcal{L}_\text{teacher}(\Generator,\GeneratorFixedScale,\LatentCode) - \frac{\lambda_{R_1}}{2} R_1(\Discriminator, \ImageSample).
\end{split}
\end{equation}
We use $\lambda_\text{teacher} = 5$; other values offer slight tradeoffs between
similarity to the base teacher model \GeneratorFixedScale, and FID score (see supplemental).
$\mathcal{L}_\text{teacher}$ is an auxiliary loss to encourage faithfulness to
the pretrained fixed-resolution generator \GeneratorFixedScale.
The architecture of \Discriminator remains the same as in the pretraining step; we found that modifying the discriminator setup did not further improve results (see supplemental).

\myparagraph{Teacher model.}
For the second training phase above, we initialize \Generator with the
pretrained weights of \GeneratorFixedScale. Weights for discriminator \Discriminator are also kept for fine-tuning.
We keep a separate copy of \GeneratorFixedScale with frozen weights, the teacher, for additional supervision.
We design a loss function that encourages the generated patch (at any
resolution), to match the teacher's fixed-resolution output in the region
corresponding to the patch, after downsampling and proper alignment\cite{irani1991improving}.
Formally, this loss is given by:
\begin{equation}\label{eqn:teacher}
\mathcal{L}_\text{teacher}(\Generator, \GeneratorFixedScale, \LatentCode) = 
\Distance\left(\Mask \odot \Warp(\Generator(\LatentCode,\Coordinates,\FullImageResolution)),
 \; \Mask \odot \GeneratorFixedScale(\LatentCode)
 \right),
\end{equation}
where \Distance is the sum of a pixel-wise
$\ell_1$ loss, and the LPIPS perceptual distance \cite{huh2020transforming,zhang2018unreasonable}.
The warp function \Warp transforms and resamples the generated high-resolution patch using a band-limited
Lanczos kernel, to project it in the coordinate frames of the low-resolution, global image $\GeneratorFixedScale(\LatentCode)$.
Because the warped patch does not cover the entire image domain, we multiply it with a binary mask \Mask to indicate the valid pixels, prior to computing loss $\Distance$.

\subsection{Implementation details}\label{sec:details}

\myparagraph{Scale-conditioning.}
In addition to the pixel location \Coordinates, we also pass the global
resolution information \FullImageResolution to the generator.
Knowledge of the global image scale is important to enable continuous scale
variations and proper anti-aliasing~\cite{chen2021learning,barron2021mip}.
We found it beneficial to explicitly inject this information into all intermediate layers of the generator.
To achieve this, we use a dual modulation approach~\cite{zhao2021comodgan},
embedding the latent code \LatentCode and scale \FullImageResolution separately
using two independent sub-networks (we use the same mapping network architecture for each). %
The two outputs are summed to obtain a set of modulation parameters
$\AuxNetwork(\LatentCode,\FullImageResolution)$, used to modulate the main
generator features.
Architectural details of the generator \Generator and mapping network \AuxNetwork,
can be found in the supplemental.

\myparagraph{Synthesizing large images.}
Our fully-convolutional generator can render image at arbitrary resolutions.
But images larger than $1024\times 1024$ require significant GPU memory.
Equivalently, we can render non-overlapping tiles that we assemble into a larger image.
Our patch-based multi-resolution training and the Fourier encoding of the
spatial coordinates make the tile junctions seamless.

\section{Experiments}\label{sec:experiment}

We introduce a modified image quality metric that computes FID over multi-scale image patches without downsampling, which is largely correlated to the standard FID metric when ground truth high-resolution images are available (FFHQ), yet more sensitive to the quality of larger resolutions.
We then compare our model to alternative approaches for variable scale synthesis and super-resolution on other natural image domains (\S~\ref{sec:expt_continuous_generation}).
Finally, we investigate variations of our model and training procedure to validate our design decisions (\S~\ref{sec:expt_model_variations}).

\begin{figure}[t!]
\centering
\includegraphics[width=\textwidth]{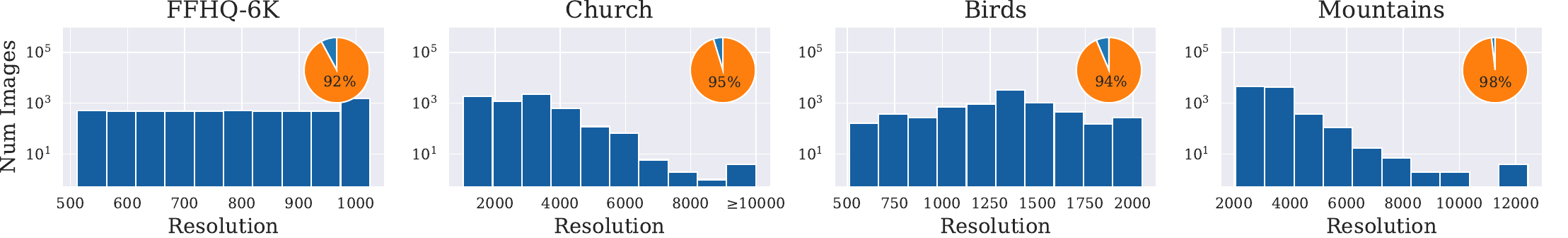}
\vspace{-4mm}
\caption{\label{fig:dataset}\small
Training set size distributions. Histogram shows the size distribution of the HR images (y-axis in log scale); pie chart indicates the proportion of LR to HR images.}
\end{figure}

\myparagraph{Data.} Our method is general and can work on collections of
any-resolution data.
As such, when targeting high-resolution generation, rather than starting over,
we can add additional high-resolution (HR) images to existing, fixed-size,
low-resolution (LR) datasets.
Our datasets and their statistics are listed in Tab.~\ref{tab:dataset}.
Figure~\ref{fig:dataset} shows the resolution distrbution in each dataset.

We begin initial experiments with a controlled setting of FFHQ, which contains
70K images at 1024 resolution. From these, we construct a varied-size dataset by
(1) using 256 resolution for all images (2) downsampling a 5K subset between
512-1024 (uniformly distributed) and (3) add 1k subset at full 1024.
The last step enables us to judiciously compare to methods that are limited to
synthesizing images at strict powers of 2. We refer to this mixture as
FFHQ6k. We use the full 1024 dataset as ground-truth for
evaluation metrics.

In the remaining domains, we push current generation results to higher resolution by scraping HR images from Flickr. In cases where a standard
fixed-size dataset is available (LSUN Churches~\cite{yu15lsun} and Mountains~\cite{park2020swapping}),
we select the additional HR images to approximately match the LR domain.
Our final generators synthesize realistic details despite the majority of the training set being LR. For Birds and Churches, $>$ 92\% of the training set is at 256 resolution but our model maintains quality beyond 1024; for Mountains $>$ 98\% of training set at 1024 but our model can generate beyond 2048.
These categories cover a range between objects (but without the strong alignment of FFHQ) and outdoor scenes.

\begin{table}[tb!]
\centering
\caption{\small Any-resolution datasets and generator settings. We build upon low-resolution (LR) datasets, and use it for fixed-size dataset pre-training. We add additional high-resolution (HR) images, of mixed resolutions. Note that the number of HR images is small ($\sim$2-8$\%$ of LR size). Patches of size $\PatchResolution$ are sampled from both subsets during training, with average sampled scale $\AvgResolution$.\label{tab:dataset}}
  \resizebox{1.0\linewidth}{!}{
  \begin{tabular}{lc@{\hskip 1mm}r@{\hskip 2mm}r@{\hskip 3mm}rrr@{\hskip 1mm}rcc@{\hskip 1mm}rcrr}
    \toprule

    \multirow{3}[2]{*}{\bf Domain} & \multicolumn{7}{c}{\bf Dataset} & \multicolumn{3}{c}{\bf Generator} \\
    \cmidrule(lr){2-8} \cmidrule(lr){9-11}

    & \multirow{2}[2]{*}{\shortstack[c]{\bf Source}} & \multicolumn{2}{c}{ $\#$ Imgs } & \multicolumn{4}{c}{ Resolutions }
    & \multirow{2}{*}{Config} & \multicolumn{2}{c}{Resolutions} \\
    \cmidrule(lr){3-4} \cmidrule{5-8} \cmidrule{10-11}

    & & \multicolumn{1}{c}{LR} & \multicolumn{1}{c}{HR} & \multicolumn{1}{c}{LR} & HR$_\text{min}$ & HR$_\text{med}$ & HR$_\text{max}$ & & $\PatchResolution$ & $\AvgResolution$ \\

    \midrule
    Faces & FFHQ & 70,000 & 6000 & 256 & 512 & 819 &  1024 & R & 256 & 458 \\ %
    Churches & LSUN \& Flickr & 126,227 & 6253 & 256 & 1024 & 2836 & 18,000 & T & 256 & 1061 \\ %
    Birds & Flickr & 112,266 & 7625 & 256 & 512 & 1365 & 2048 & T & 256 & 585 \\% 1024 \\
    Mountains & Flickr & 507,495 & 9361 & 1024 & 2049 & 3168 &  12,431 & T & 1024 & 1823 \\ %
    \bottomrule
  \end{tabular}
  }
\end{table}

\subsection{Continuous Multi-Scale Image Synthesis.}\label{sec:expt_continuous_generation}

\begin{figure}[p!]
\centering
\includegraphics[width=\textwidth]{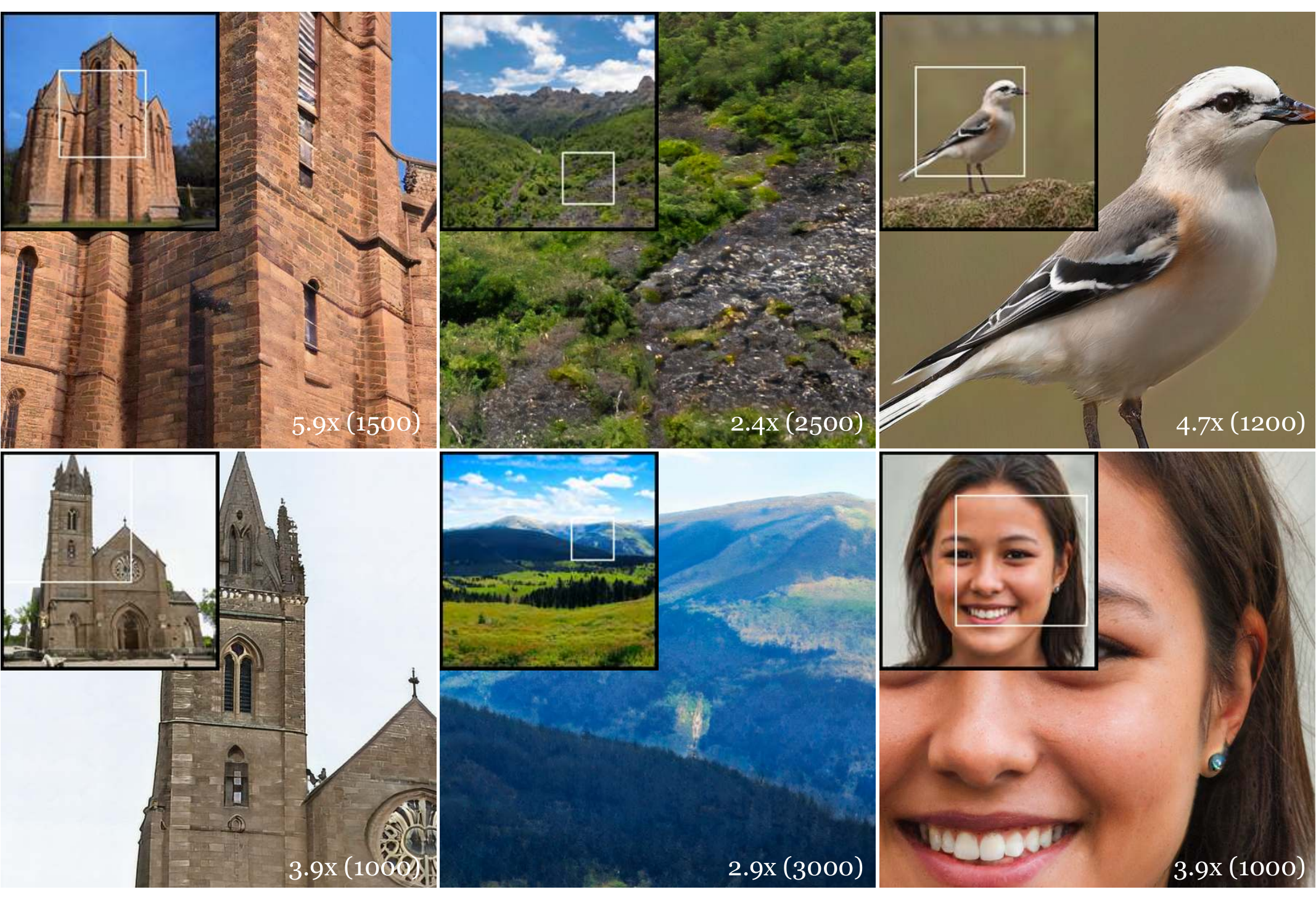}
\vspace{-7mm}
\caption{\label{fig:qualitative}\small
The inset shows the entire generated, high-resolution images (between 1000-3000
resolution), with enlarged regions of interest outlined in the white box. Note
that our model can render the image (or any sub-region) at any resolution.
}
\end{figure}

\myparagraph{Qualitative examples.} We show qualitative examples in
Fig.~\ref{fig:qualitative}.
Our generated images preserve the fine details of
HR structures, such as bricks, rocky slopes, feathers, or hair.
Pushing the inference resolution towards and beyond the higher resolutions of training images, we find that textured surfaces typically deteriorate first before edge boundaries deteriorate eventually (Fig.~\ref{fig:extrapolation}).

\myparagraph{Patch-FID metric.} 
Standard FID evaluates \textit{global structure} by first downsampling all images to a common size of 299. By design, this ignores high-resolution details (and itself can cause artifacts~\cite{parmar2021cleanfid}). 
Therefore, we propose a modification, which we dub `patch-FID', to specifically evaluate \textit{texture} synthesized at higher resolutions. 
Our patch-FID randomly resizes and crops patches from the HR dataset, and computes FID on real and generated patches, sampled at corresponding scales and locations. We use 50k patches, matching standard FID.
By avoiding downsampling, our patch-FID is more sensitive to blurriness or artifacts at higher resolutions, resulting in larger absolute difference compared to standard FID at 1024 resolution. As a sanity check, when a full HR ground-truth is available, we find it is largely correlated to standard FID (see Table~\ref{tab:anycost}).

To summarize, to evaluate structure, we compute standard FID on images generated at specified resolutions, e.g., FID (256). To evaluate texture, we sample patches at random scales and locations and measure our patch-FID, which we denote as pFID (random). Lower numbers are better in both cases.

\begin{figure}[p!]
\centering
\includegraphics[width=\textwidth]{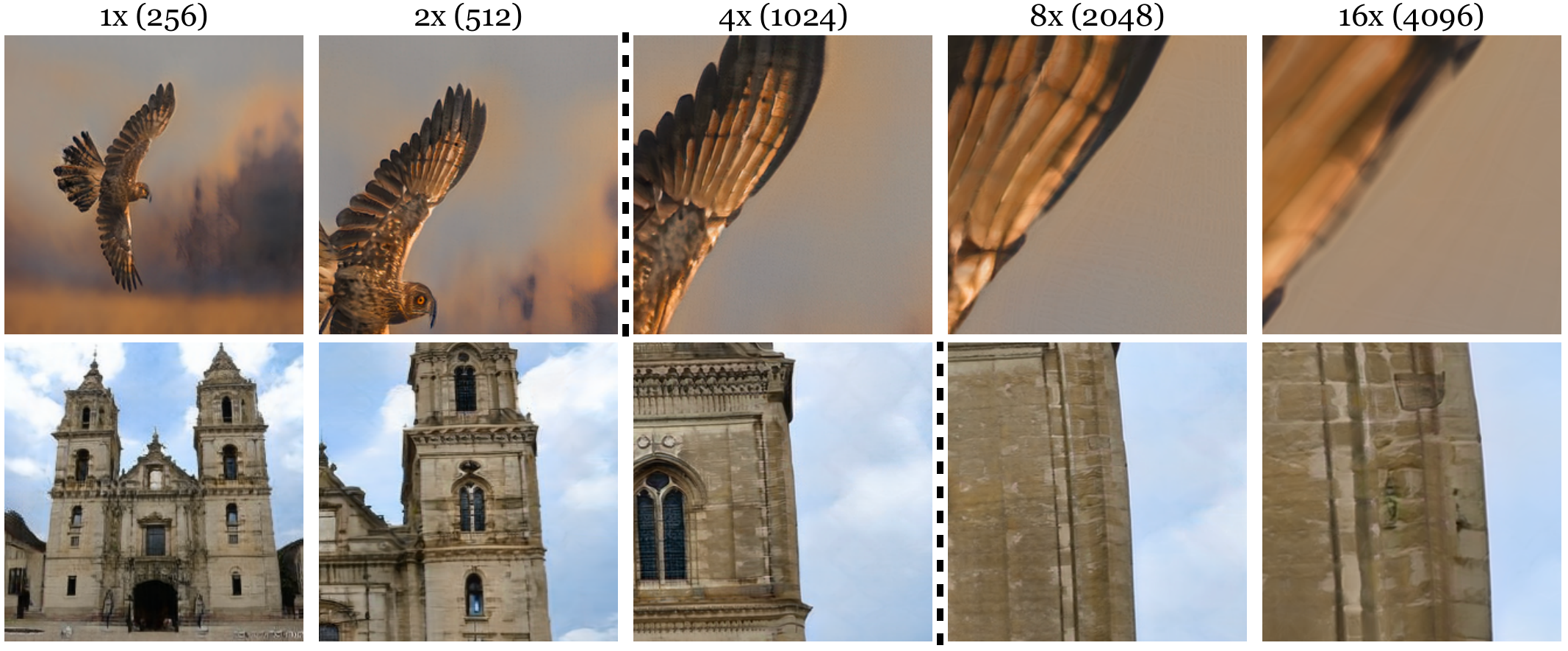}
\caption{\label{fig:extrapolation}\small Extrapolation limits. We test the
extrapolation capabilities of our model by specifying the inference scale
\FullImageResolution. Typically,
textures such as bricks and feathers deteriorate first before edges degrade.
The dotted line indicates when generation starts to exceed the average
scale sampled in training \AvgResolution (which is 585 for birds and 1061 for churches).}
\end{figure}

\myparagraph{Alternative methods of multi-size training and generation.} Our generator is encouraged to synthesize realistic high-resolution textures at training, even when the discriminator does not get to see the full image. %
While MS-PE~\cite{choi2021toward} also enables continuous resolution synthesis, the discriminator learns only at a single resolution and the generator is not trained patch-wise. We find that this downsampling for the discriminator is detrimental to image quality at higher resolutions. Anycost-GAN\cite{lin2021anycost} performs image synthesis at powers-of-two resolutions by adding additional synthesis blocks. For comparison, we modify it to handle a multi-size dataset by downsampling images to the nearest power of two and training each layer only on the valid image subset. Compared to Anycost-GAN, our model is more data-efficient, due to weight sharing for generation at multiple scales. Anycost-GAN learns a separate module for each increase in resolution, creating artifacts at higher resolutions when fewer HR training images are available and higher FID scores (Tab.~\ref{tab:anycost}). Additionally, Anycost-GAN increases the generator and discriminator size for synthesis at higher resolutions, whereas our model incurs a constant training cost, regardless of the inference scale.

\begin{table}[t!]
\parbox[t]{.46\linewidth}{
\caption{\small Varied-size training and inference.
Random-resize MS-PE~\cite{choi2021toward} performs varied-size synthesis, but assumes a fixed-size dataset. AnyCost-GAN handles varied training at powers of 2. Our method
directly utilizes training images at any size, achieving better results by FID. ($\dagger$ = copied from paper)\label{tab:anycost}}
\centering
    \resizebox{0.82\linewidth}{!}{
  \begin{tabular}{lcccc}
    \toprule

    \multirow{2}{*}{FFHQ6K} & \multicolumn{3}{c}{FID} & pFID \\
    \cmidrule(lr){2-4}\cmidrule{5-5}
     & 256 & 512 & 1024 & random \\

    \midrule
    MS-PE~\cite{choi2021toward}$\dagger$         & 6.75 & 30.41 & -- & -- \\   
    Anycost~\cite{lin2021anycost}        & 4.24 & 5.94 & 6.47 & 18.39 \\
    Ours & 3.34 & 3.71 & 4.06 & 2.96 \\
    \bottomrule
  \end{tabular}
  }
}
\hfill
\parbox[t]{.51\linewidth}{
\caption{\small Comparison to super-resolution using patch-FID (pFID). For each domain, we compare our model to continuous-scale (LIIF) and fixed-scale super-resolution (Real-ESRGAN) models.
Lower pFID suggests that our model can generate realistic details at high resolutions, not achievable with super-resolution alone. \label{tab:SR}}
\centering
  \resizebox{1.05\linewidth}{!}{
  \begin{tabular}{lcccc}
    \toprule
        & \multicolumn{4}{c}{pFID (random)} \\
        \cmidrule(lr){2-5}
                & FFHQ6K & Church & Bird & Mountain \\
    \midrule
    LIIF~\cite{chen2021learning}             & 22.93 & 83.88 & 30.19 & 23.10 \\
    Real-ESRGAN~\cite{wang2021real}           & 16.92 & 23.04 & 16.10 & 19.05 \\
    Ours           & 2.96 & 9.89 & 6.52 & 7.99 \\
    \bottomrule
  \end{tabular}
  }
}
\vspace{-5mm}
\end{table}

\begin{figure}[!tb]
\centering
\includegraphics[width=\textwidth]{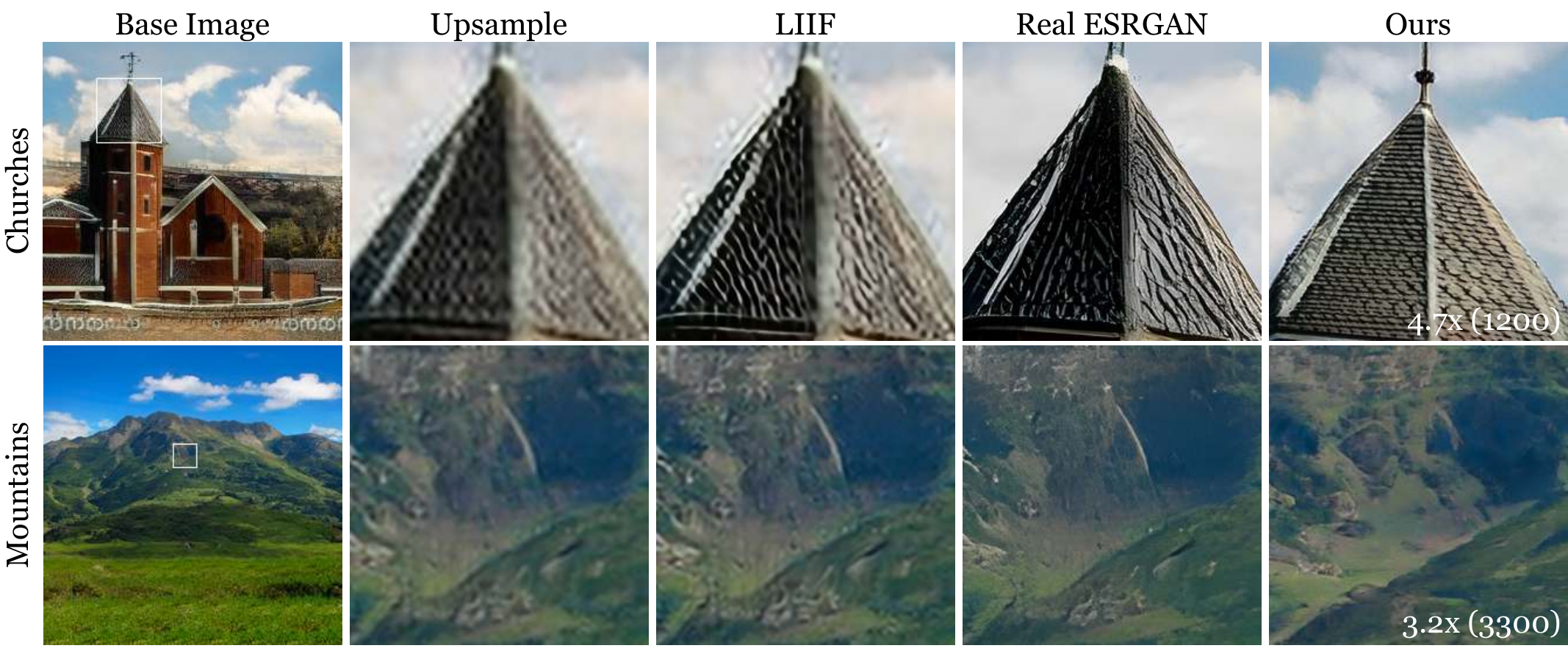}
\caption{\small Super-resolution Comparisons. Qualitative comparisons of Lanczos upsampling a patch from the base image (upsample), continuous (LIIF~\cite{chen2021learning}) and fixed-factor (Real ESRGAN~\cite{wang2021real}) super-resolution models, and our trained model. LIIF tends to amplify artifacts from the base image (e.g. the JPEG artifacts around the church). While Real-ESRGAN is better at suppressing artifacts, it tends to overly smooth surfaces or synthesize grid-like textures (mountain). Our model is not a super-resolution model; it can add additional details to the low-resolution image but tolerates slight distortions in structure which are regularized with the teacher weight.}
\label{fig:superresolution}
\vspace{-2mm}
\end{figure}

\myparagraph{Comparison to super-resolution.}
Most super-resolution methods require LR/HR image pairs, whereas there is no
ground-truth HR counterpart to a LR image synthesized by our fixed-scale
generator \GeneratorFixedScale.
The teacher regularization encourages similarity between \GeneratorFixedScale
and \Generator's outputs, but unlike super-resolution, this supervision occurs
at low-resolution, allowing variations in fine details.
Figure~\ref{fig:superresolution} compares our model to super-resolution
methods applied to the output of \GeneratorFixedScale.
Our method produces much sharper details than LIIF~\cite{chen2021learning},
a recent continuous-scale super-resolution technique, and cleaner images
than the state-of-the-art Real-ESRGAN~\cite{wang2021real}.
The latter is a fixed-resolution model, so we run it iteratively until exceeding a target resolution, and then Lanczos downsample the result to the target size.
Real-ESRGAN's outputs are either overly smooth, or exhibit grid-like artifacts.
Our method generates realistic textures based on the
low-resolution output of \GeneratorFixedScale and reaches a lower pFID
(Tab.~\ref{tab:SR}). 

\subsection{Model Variations}\label{sec:expt_model_variations}

Using the full high-resolution FFHQ dataset as a benchmark, we investigate
individual components of our architecture and training process.
We train each model variation for 5M images and record metrics from the best
FID@1024 checkpoint. 
We only report quantitative metrics in the main paper and refer to the supplemental for further
evaluations and visual comparisons.

\myparagraph{Impact of teacher regularization.}
Our full model uses an ``inverse'' teacher regularizer to
encourages a downsampled HR patch to match the low-resolution
teacher as described in~\ref{sec:training}.
We also explored a variant with a ``forward'' teacher loss, in which the
generated patch is encouraged to match the \emph{upsampled} teacher output.
This variation is qualitatively inferior and blurs details; it has worse FID at higher resolutions
(see supplemental for details and visuals).
Removing the teacher altogether improves pFID but degrades FID.
Qualitatively, the
generated patches diverge significantly from the fixed-size global image.
We hypothesize that the global
change in structure negatively impacts overall image quality, causing global
FID metrics to increase, but this cannot be captured from evaluating patches alone.
We found $\lambda_\text{teacher}=5$ to provide the best balance between global and local
image quality, but we observe minimal differences in FID and pFID for other values, evidence
that the model can tolerate a range of values for this parameter.
See supplemental for a parameter sweep with full scores.

\myparagraph{Removing scale conditioning degrades quality.}
We inject the scale information to intermediate layers of the generator
through scale-conditioning.
Adding this improves FID@1024 from 4.88 to 4.47, and pFID from 4.67 to 4.28.

\begin{table}[t]
\vspace{-5mm}
\parbox{.48\linewidth}{
\caption{ \small Multisize Training. Downsampling or upsampling all images to a common size, or using only the subset of the largest images, worsens FID compared to our training strategy. (*) indicates our default setting. \label{tab:multisize}}
\centering
    \resizebox{1.05\linewidth}{!}{
  \begin{tabular}{lcccc}
    \toprule
    & \multicolumn{3}{c}{FID} &pFID \\
    \cmidrule(lr){2-4}\cmidrule(lr){5-5} 
    & 256 & 512 & 1024 & random \\

    \midrule
    Resize down to 512 & 3.31 & 4.11 & 19.18 & 26.83 \\
    Resize up to 1024 & 3.46 & 13.43 & 4.86 & 6.65 \\
    Train 1024 subset & 3.46 & 12.41 & 4.67 & 5.43 \\
    Multisize training (*) & 3.37 & 4.41 & 4.47 & 4.28 \\
    \bottomrule
  \end{tabular}
  }
}
\hfill
\parbox{.48\linewidth}{
\caption{\label{table:hr_images}\small Number of HR images. Our method is robust to a wide range of HR
images, even when only 1K images at HR are available ($<$2\% of the full ground-truth dataset). 
\label{tab:num_images}}
\centering
  \resizebox{0.95\linewidth}{!}{
  \begin{tabular}{lcccc}
    \toprule
    & \multicolumn{3}{c}{FID} &pFID \\
    \cmidrule(lr){2-4}\cmidrule(lr){5-5} 
    & 256 & 512 & 1024 & random \\

    \midrule
    1k  (1.4\%)            & 3.43 & 5.13 & 4.38 & 3.73 \\
    5k  (7.1\%)  (*)           & 3.36 & 4.97 & 4.54 & 3.48 \\
    10k  (14.2\%)           & 3.46 & 4.96 & 4.65 & 3.54 \\
    70k (100\%)            & 3.42 & 4.88 & 4.52 & 3.42 \\
    \bottomrule
  \end{tabular}
  }
}
\vspace{-5mm}
\end{table}

\myparagraph{Multi-size training improves fidelity at all scales.}
Our multi-size data pipeline lets our model learn to synthesize at continuous
scales, which is a strictly more challenging than learning at a fixed scale.
In Table~\ref{tab:multisize}, we investigate to what extent learning from images of varied sizes offers benefits
over fixed-scale training on a smaller dataset.
Visual comparisons can be found in supplemental.
In a first alternative, using the same FFHQ-6K dataset, we resize all images down to 512 and train the model to generate patches at $512\times 512$.
In this case, the model
performs well up to 512 scale, but does not generalize beyond (e.g., 1024) since it cannot 
exploit the information lost in downsampling.
In two other variants, we train models for 1024 resolution
in the first case by upsampling all images up to 1024, in the second by keeping only the 1K subset of images at 1024 resolution.
Both variants are trained to output images specifically at 1024 resolution.
The former approach (upsampling) increases blurriness.
In the latter, FID@1024 remains worse than that of our
multi-size training, which can take advantage of more data despite most of it
being \emph{smaller} than 1024.

\myparagraph{Impact of number of HR images}
Due to the patch-based
training procedure, we find that our model can be trained with a small fraction of HR
images, compared to the 70k LR images in the dataset.
In Table \ref{tab:num_images}, we use progressively larger subsets of HR images: 1k, 5k, 10k.
We found that the FID scores are largely
similar (within 0.3) to using the entire 70k HR images. %
However, training with 1k or fewer HR images shows evidence of divergence during training (see supplemental),
but stabilizes by 5k HR images, For the remaining domains, we collect roughly 5K-10K images to construct the HR dataset.

\subsection{Properties of Multi-Scale Generation}\label{sec:expt_properties}

\begin{figure}[!bt]
\centering
\includegraphics[width=\textwidth]{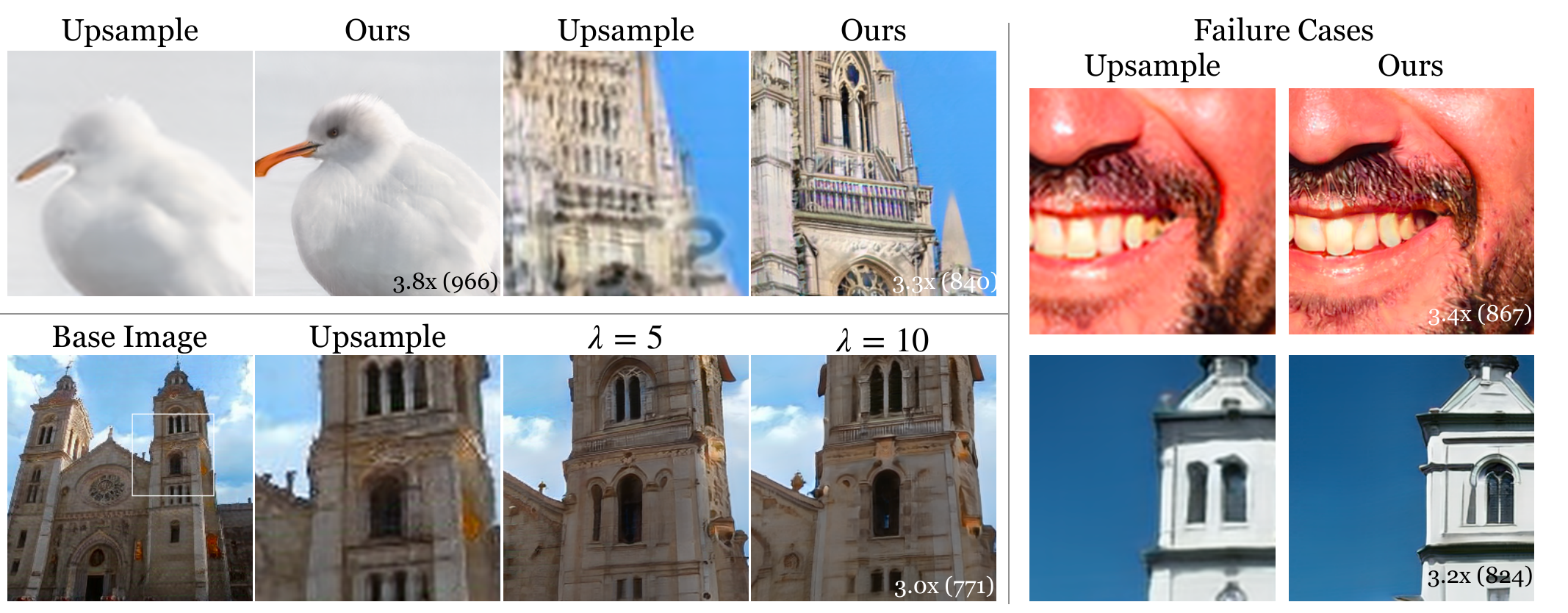}
\vspace{-7mm}
\caption{\small Model Properties and Failure Cases. As fine details can be more difficult to learn at low resolution, our model is capable of adding corrections when generating at higher resolutions. In the case of inconsistencies between the LR and HR data sources, the model deletes patterns that are not present in the HR dataset (\eg watermarks and compression artifacts), influenced by the teacher regularization weight. Failure cases include biases towards circular or ring-like structures.}
\label{fig:properties}
\vspace{-5mm}
\end{figure}

\myparagraph{Correcting artifacts from low resolution.}
Because our model is not
directly trained with corresponding LR and HR image pairs, we find that there
can be small distortions between the upsampled base image and the HR generation
from the same latent code.
In some cases, this can be a desirable property (Fig.~\ref{fig:properties}).
For instance, the base generator on the birds dataset
can struggle in synthesizing the eye of the bird, which is less apparent at low
resolutions, but more salient at high resolution.
Consequently, our HR generation will add the missing eye, and also
synthesizes additional feather and beak details.
In the churches domain, because
the LR and HR datasets are collected separately, we find that the synthesized
watermarks and JPEG artifacts at the base resolution disappear at higher resolution, because the HR dataset we used is of higher quality and does not have any watermark.
The similarity between the LR and HR generations can be tuned
using $\lambda_\text{teacher}$ during training. 

\myparagraph{Failure Cases.} Our model tends to inherit the artifacts from
StyleGAN3, such as a centered front tooth in FFHQ.
In instances in which the base resolution image contains uneven
surfaces, the model may fail to fully mitigate them at higher resolutions. These
artifacts are often subtle at the low resolution, but become more apparent when
upsampling the base image or generating at a larger target scale. In some cases,
our model also has a tendency to generate ``watery'' circular or ring-like artifacts
(Fig.~\ref{fig:properties}).

\section{Conclusion}

We propose an image synthesis approach that can train on images of varied resolution and perform inference at continuous resolutions.
This lifts the fixed-resolution requirement of prior generative models, which discard higher-resolution details.
To do this, we train a generator jointly on a low-resolution dataset 
to learn global structure, and on patches from the varied-size dataset to learn details.
At inference time, we can synthesize an image at any resolution by supplying the appropriate coordinate grid and scale factor to the generator.
By using training images at their native resolutions and a single model for continuous-resolution synthesis, our method can efficiently leverage information present in only a handful of high-resolution images to
complement a large set of low-resolution images.
This approach enables high-resolution synthesis without a larger generator or large dataset of fixed-size, high-resolution images.

\subsubsection{Acknowledgements.} We thank Assaf Shocher for feedback and Taesung Park for dataset collection advice. LC is supported by the NSF Graduate Research Fellowship under Grant No. 1745302 and Adobe Research Fellowship. This work was started while LC was an intern at Adobe Research and was supported in part by an Adobe gift.

\bibliographystyle{splncs04}
\bibliography{egbib}

\newpage
\section*{Supplementary}

In the supplementary, we first demonstrate an extension of our approach towards panorama generation (Section~\ref{sec:sm_panorama}). In Section~\ref{sec:sm_experiments}, we provide additional qualitative and quantitative comparisons to baselines (super-resolution methods, discrete-resolution and oracle generators), explore additional model variations, and investigate the detectability of our method using an off-the-shelf forensics method. We provide implementation details in Section~\ref{sec:sm_method}.
\section{Panorama generation extension}
\label{sec:sm_panorama}

Our default training setup assumes that we use low-resolution images to learn global context and patches from high-resolution images to learn details. An alternative setup to learn from patches directly, \textit{without ever knowing the entire global context}. One such scenario is panorama generation, where a large-scale dataset would be much more difficult to obtain than single images. We investigate this setup on the Mountains domain, in which the generator is tasked with synthesizing a panorama from landscape images, without training directly on panoramas. Accordingly, we modify the \ContinuousDomain coordinate grid to $[-\pi,\pi]\times[0,1]$, and enforce continuity on the endpoints by using a sine and cosine encoding prior to Fourier feature embedding. At training time, we sample a ``slice'' of the coordinate grid for generation corresponding to a random viewing angle, but at inference time the entire panorama can be synthesized by specifying the full grid of coordinates. In this case, we find that it is important to use a \textit{cross-frame} discriminator, in which the discriminator straddles the boundary between two generated slices to enable seamless boundaries in the panorama. Qualitative results are shown in Fig~\ref{fig:sm_panorama}. At inference time, we can spatially interpolate the \StyleCode latent code with arbitrary spacing, which generates seamless infinite landscapes.

\newpage

\begin{figure}[H]
\centering
\includegraphics[width=\textwidth]{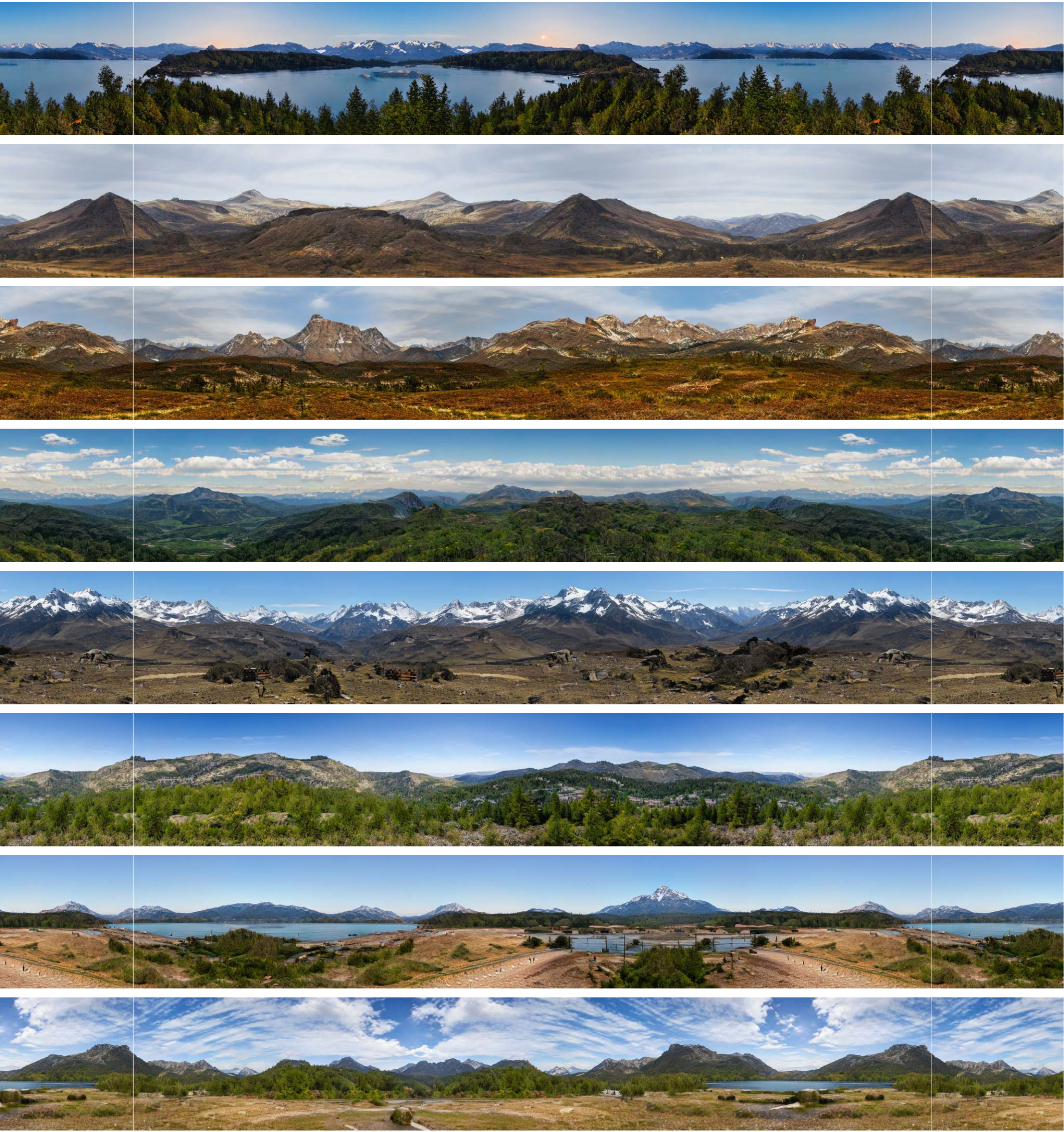}
\caption{\small Panorama generation from patches. We modify our training framework to train without the global image context. We map our coordinate grid to $[-\pi,\pi]\times[0,1]$ and use a cross-frame discriminator to enable seamless transitions between patches. The model is trained with $\mathrm{FOV}=60^{\circ}$. The vertical white line indicates a full $360^{\circ}$ revolution.}
\label{fig:sm_panorama}
\end{figure}

\begin{figure}[H]
\centering
\includegraphics[width=\textwidth]{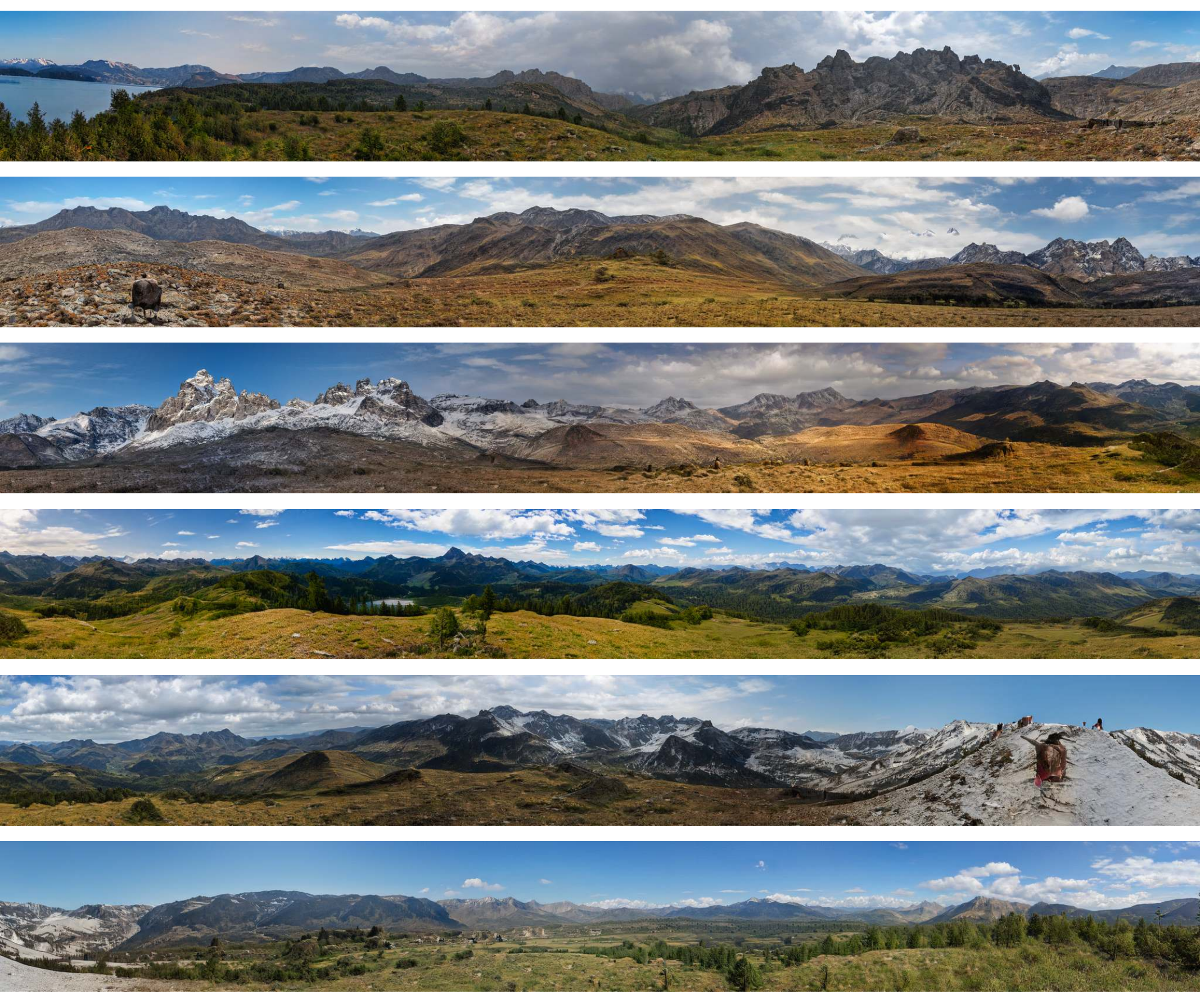}
\caption{\small By spatially adjusting the latent code only at inference time, the same model is capable of generating infinite landscapes, shown on multiple lines here for visibility.}
\label{fig:sm_panorama}
\end{figure}

\section{Experiments}\label{sec:sm_experiments}

\subsection{Dataset Collection}

\begin{wraptable}{r}{4.5cm}
\vspace{-8mm}
\caption{\small Image sources for construction of our varied-resolution datasets. \label{tab:sm_dataset_source}}
\resizebox{1.0\linewidth}{!}{
\begin{tabular}{lc}
    \toprule
    Domain & Flickr Source \\
    \midrule
    Church & Church Exteriors \\
    Mountains & Mountains Anywhere \\
    Birds & Birding in the Wild \\
    \bottomrule
  \end{tabular}
  }
\vspace{-6mm}
\end{wraptable}

To collect our varied-size dataset, we scrape image collections from Flickr photo groups (Tab.~\ref{tab:sm_dataset_source}). In cases where a standard fixed-resolution dataset is available (\eg LSUN Churches~\cite{yu15lsun}), we seek to find photos that approximately match the domain of the standard dataset. Due to domain mismatches between LSUN and the photos scraped from Flickr, we manually filter the collected images to approximately match the LSUN domain, which remains tractable for the few thousand HR images used in the patch-based training phase. As is standard practice~\cite{thomee2016yfcc100m,park2020swapping},
and to not violate license permissions, we will release the image IDs but not the images directly.

\begin{figure}[t] %
  \begin{center}
    \includegraphics[width=0.8\textwidth]{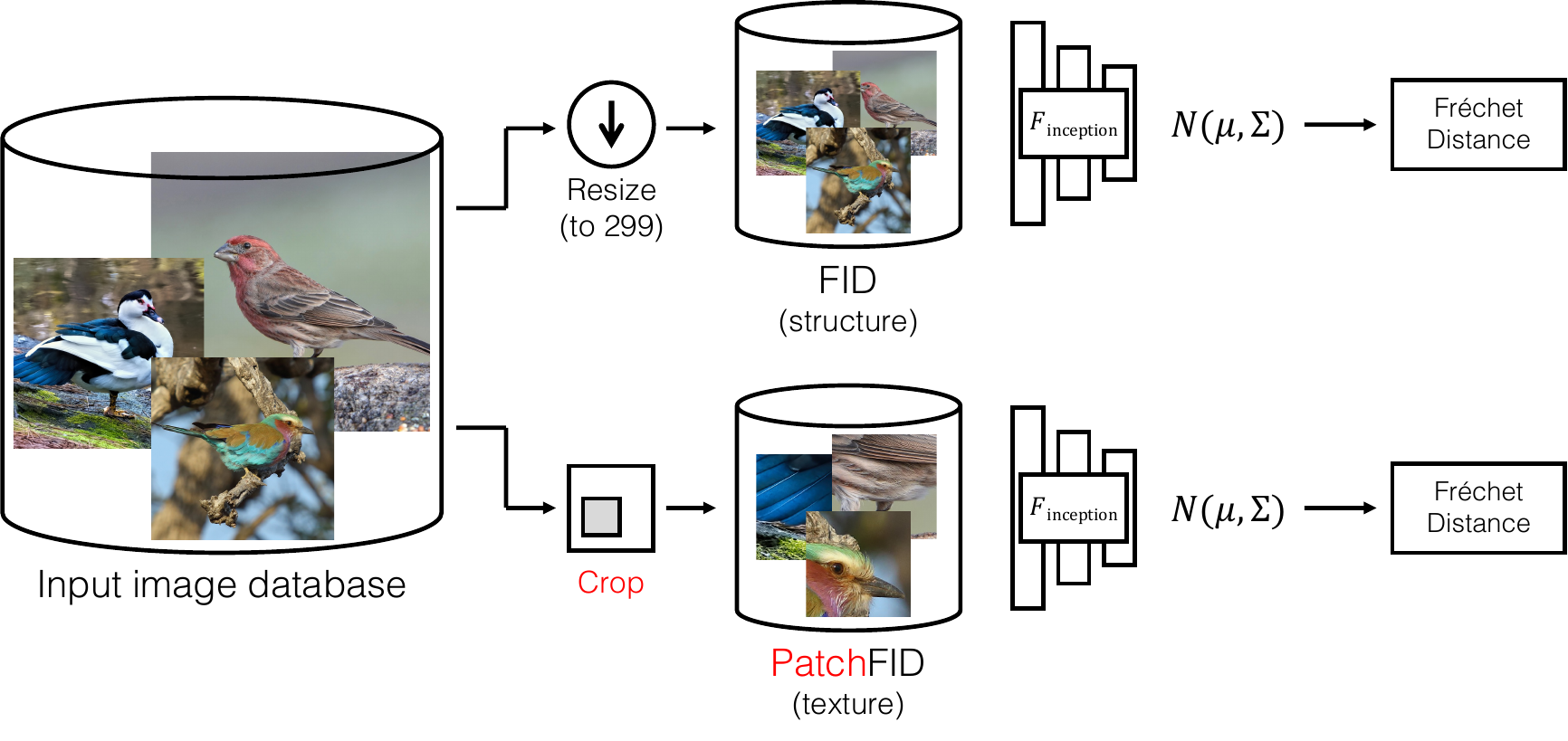}
  \end{center}
  \caption{\small Preprocessing for FID and patch-FID (pFID). FID evaluates global structure, but downsamples all images to a common 299px size which ignores higher resolution details. pFID takes images crops instead rather than downsampling to capture texture realism at higher resolutions. We use both to measure structure and texture properties. \label{fig:fid_vs_patchfid}}
\end{figure}

\subsection{Patch-FID}

\begin{wrapfigure}{r}{0.35\textwidth}
    \vspace{-2mm}
  \begin{center}
    \includegraphics[width=.35\textwidth]{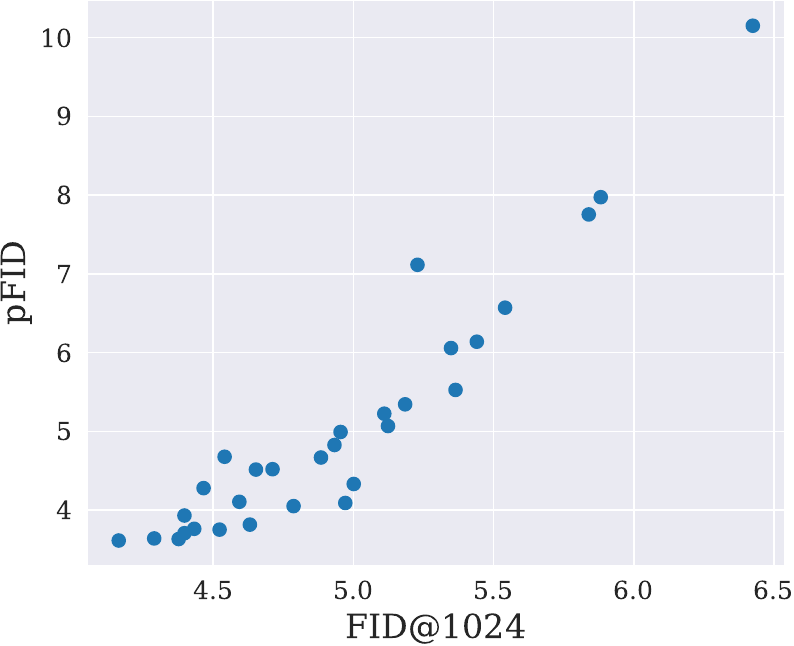}
  \end{center}
  \vspace{-4mm}
  \caption{\small pFID vs FID@1024. The metrics are largely correlated, but
  FID@1024 generates images at 1024 resolution and then downsamples images, assuming a fixed-size dataset. On other domains
  where a ground-truth HR dataset is not available, we primarily use pFID to
  measure sample quality. \label{fig:sm_correlation}}
  \vspace{-2mm}
\end{wrapfigure}

We describe details of our Patch FID metric, introduced in Section~\ref{sec:expt_continuous_generation} of the main paper. The metric is aimed at better capturing the realism of details at high resolution by avoiding downsampling to capture texture details. 
In our setting, we have a smaller number of real images present at various resolutions. Standard FID, which focuses on global structure, does not capture these varied-resolution details.
We modify the FID 
pipeline to avoid downsampling global images at higher resolutions to a fixed
299 pixel width. Instead, we randomly sample patches of size \PatchResolution from real
images at global scale \FullImageResolution and locations \Coordinates, and
generate the corresponding patch  
\Generator(\LatentCode,\Coordinates,\FullImageResolution). 
The number of samples is crucial to getting an accurate estimate in FID calculation~\cite{chong2020effectively}, and 50,000 is typically used. A benefit of this patch-sampling procedure also means that we can obtain the necessary large number of patches for FID computation, more than the available number of real images in the HR dataset.
In the Mountains
generator, where the patch size \PatchResolution is larger than 299, we
subsequently also select a random 299-pixel crop from the patches to compute the
image features. Because this avoids downsampling the generated content, we find
that it is more sensitive to image quality at high resolutions. Using the full
FFHQ dataset as ground-truth, we find that our patch-FID metric is largely
correlated to the standard FID numbers at 1024 resolution
(Fig.~\ref{fig:sm_correlation}). Therefore, we use Patch-FID as a metric of
sample quality on our datasets collected from Flickr, when a full
high-resolution dataset of images all at the same resolution is not available.

\subsection{Additional quantitative results}\label{sec:sm_results}

In Table~\ref{tab:SR} of the main text, we report
comparisons of our method and off-the-shelf super-resolution methods using the
patch-FID metric. We report additional metrics in Table~\ref{tab:sm_SR} here,
including the FID at base resolution (the result of the pretraining step), and
FID at a higher resolution after downsampling all images in the HR dataset
(between 5k-10k images, which is lower than the typical 50k used to compute FID)
to a common size.  Notably, the base resolution FID is largely similar before
and after patch-based training, and in the case of FFHQ and Birds, patch-based
training at higher resolutions even improves the low-resolution FID. Without
direct multi-scale training, however, the fixed-size model obtained from the
pretraining step does not naturally generalize to higher resolutions. We find
that our pFID metric is more discriminative to differences in image quality at
higher resolutions. In particular, the LIIF super-resolution model tends to
obtain better FID@1024 (corresponding to super-resolving generated images to 1024 resolution and computing FID) compared to Real-ESRGAN, but the outputs are visually
blurry. Because our pFID does not perform downsampling, it can better capture
this blurriness, reflected in an increased pFID. In another variant, we compute
pFID on patches of size 1024 synthesized by the Mountain generator, and then
subsequently downsample them, which we denote as pFID~(down), rather than cropping.
Again, we find that this downsampling operation can obscure image deterioration
at higher resolution, producing artificially lower FID scores compared to pFID
computed without downsampling. 

\begin{table}[t!]
\caption{\small Alternative FID evaluation metrics. We primarily use pFID, which avoids downsampling synthesized content and evaluates multiscale patches, as an evaluation metric. Here, we also report FID at the base resolution from the result of fixed-size pretraining (Fixed-Size), and compare to global FID metrics after downsampling the HR images to a common size. On FFHQ, we also tried applying GFP-GAN~\cite{wang2021gfpgan} which is a facial super-resolution model.  \label{tab:sm_SR}}
\centering

\resizebox{1.0\linewidth}{!}{
  \begin{tabular}{lccc@{\hskip 2mm}@{\hskip 2mm}ccc @{\hskip 2mm}@{\hskip 2mm}ccc @{\hskip 2mm}@{\hskip 2mm}ccc}
    \toprule

    & \multicolumn{3}{c@{\hskip 2mm}@{\hskip 2mm}}{\bf FFHQ6K} & \multicolumn{3}{c@{\hskip 2mm}@{\hskip 2mm}}{\bf Church} & \multicolumn{3}{c@{\hskip 2mm}@{\hskip 2mm}}{\bf Birds} & \multicolumn{3}{c}{\bf Mountain} \\

    \cmidrule(lr){2-4}
    \cmidrule(lr){5-7}
    \cmidrule(lr){8-10}
    \cmidrule(lr){11-13}
    
    & \multicolumn{2}{c}{FID} & pFID
    & \multicolumn{2}{c}{FID} & pFID 
    & \multicolumn{2}{c}{FID} & pFID 
    & FID & \multicolumn{2}{c}{pFID}  \\
    
    \cmidrule(lr){2-3} \cmidrule(lr){4-4} 
    \cmidrule(lr){5-6} \cmidrule(lr){7-7}
     \cmidrule(lr){8-9} \cmidrule(lr){10-10}
     \cmidrule(lr){11-11} \cmidrule(lr){12-13}
    & 256 & 1024 & random & 256 & 1024 & random & 256 & 512 & random & 1024 & down & random \\
    \midrule
    Fixed-Size            & 3.71 & 33.80 & 52.95 & 3.39 & 242.10 & 146.24 & 3.92 & 12.69 & 55.42 & 3.09 & 13.42 & 46.20 \\
    Upsample     & -- & 11.70 & 17.29 & -- & 14.21 & 80.48 & -- & 7.67 & 30.57 & -- & 4.53 & 20.00 \\
    LIIF            & -- & 7.05 & 22.93 & -- & 18.66 & 83.88 &  -- & 7.29 & 30.19 & -- & 4.55 & 23.10  \\
    Real-ESRGAN           & -- & 19.04 & 16.92 & -- & 12.26 & 23.04 & -- & 8.51 & 16.10 & -- & 7.60 & 19.05\\
    GFP-GAN          & -- & 19.15 & 16.27 & -- & -- & -- & -- & -- & -- & -- & -- & -- \\
    Ours           & 3.34 & 4.06 & 2.96 & 3.84 & 6.98 & 9.89 & 3.78 & 6.29 & 6.52 & 3.14 & 4.33 & 7.99  \\

    \bottomrule
  \end{tabular}
  }

\vspace{-5mm}
\end{table}

\subsection{Comparison to powers-of-two synthesis} 
\begin{figure}[ht!]
\centering
\includegraphics[width=\textwidth]{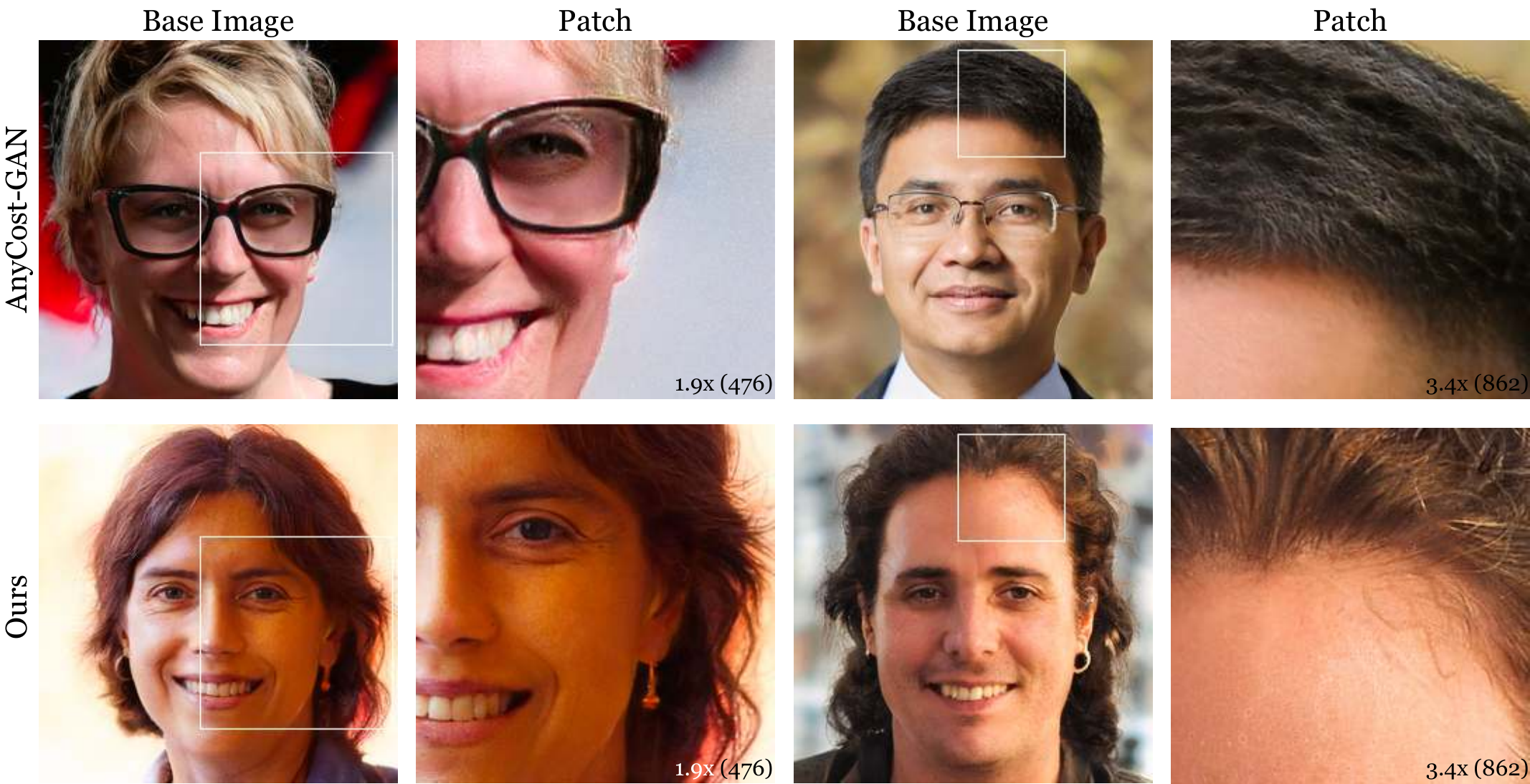}
\caption{\label{fig:sm_anycost}\small Using the same FFHQ6K dataset, we train an Anycost-GAN~\cite{lin2021anycost} and compare it to our model. While Anycost-GAN adds additional modules to increase synthesis resolution, our model shares weights across resolutions. Note that the output from Anycost-GAN contain more visual artifacts, particularly in finely textured regions such as hair.}
\end{figure}

Using the same set of
generator weights, our model can synthesize images at a specified scale
by simply providing the corresponding \FullImageResolution and \Coordinates
inputs. On the other hand, other methods for multi-resolution
synthesis~\cite{lin2021anycost,karras2017progressive,denton2015deep,karnewar2020msg}
generate images that are iteratively enlarged by a factor of two, by adding
additional network layers. These methods are typically introduced to impove
training stability. For this baseline, we modify the recent Anycost-GAN~\cite{lin2021anycost}
framework to fit our varied-size training setting. Specifically, we downsample
all images in FFHQ6K to the nearest power of two, and train the corresponding
network layers only on the appropriate subset of data. To generate at any
resolution below 1024, we take the nearest model output that is larger than the
target resolution, and apply Lanczos downsampling. Similar to our approach, we
start with a pretrained model at 256 resolution, and initialize both the
generator and discriminator with pretrained weights. Because each increase in
output resolution involves training additional weights, and the number of images
at a given resolution decreases as resolution increases, we find that this
training approach yields visual artifacts at higher resolutions, shown in
Fig.~\ref{fig:sm_anycost}. 

\subsection{Comparison to Oracle Generator}\label{sec:sm_oracle}

In Section~\ref{sec:experiment} of the paper, we describe our experimental setup on the face domain, and in Table~\ref{tab:num_images}, we show competitive performance training on few HR images, even compared training on the whole HR dataset. We provide additional details and visualizations here.

We use the FFHQ dataset as a collection of 70k high-resolution
ground-truth images. To simulate more ``in-the-wild''
settings, we use a fraction (6k) of HR images for patch-based training
with a generator of size $\PatchResolution=256$, where only 1k of the images are
the full 1024 resolution, and the remainder are uniformly downsampled between
512 and 1024 prior to training. As a comparison, we also evaluate two oracle
models that train a generator directly for the $s=1024$ global image, using the
entire 70K images in the FFHQ dataset, and Lanczos downsample the result for FID
computations at other resolutions.
We also evaluate a variant of pFID, by holding the scale \textit{fixed} at the maximal resolution and randomly sampling crop locations (pFID 1024), in addition to our original pFID metric that randomly samples both scale and location (pFID random). Note that this evaluation is only possible for FFHQ controlled setting, as all images are present at the maximal 1024 size.

\begin{table}[t]
\parbox{1.0\linewidth}{
\caption{ \small Comparison of our patch generator (6k images, varied sizes) to oracle generators which train on the entire FFHQ dataset (70k images, 1024 resolution). Although the oracle generators attain better FID, our method enables synthesis at continuous resolutions and can train without assuming that all images are resized to a common resolution. We also include pFID at the maximum 1024 scale. 
\label{tab:sm_oracle}}
\vspace{2mm}

\centering
  \resizebox{0.5\linewidth}{!}{
  \begin{tabular}{lccccc}
    \toprule
    & \multicolumn{3}{c}{FID} & \multicolumn{2}{c}{pFID} \\
    \cmidrule(lr){2-4}\cmidrule(lr){5-6} 
    & 256 & 512 & 1024 & random & 1024 \\

    \midrule
    SGAN2 Oracle           & 3.05 & 2.81 & 2.69 & 2.26 & 4.83 \\
    SGAN3 Oracle          & 3.54 & 3.23 & 3.06 & 2.44  & 4.29 \\
    Patch (Ours)          & 3.34 & 3.71 & 4.06 & 2.96 & 8.01 \\
    \bottomrule
  \end{tabular}
  }
}
\vspace{-4mm}
\end{table}

\begin{figure}[ht!]
\centering
\includegraphics[width=\textwidth]{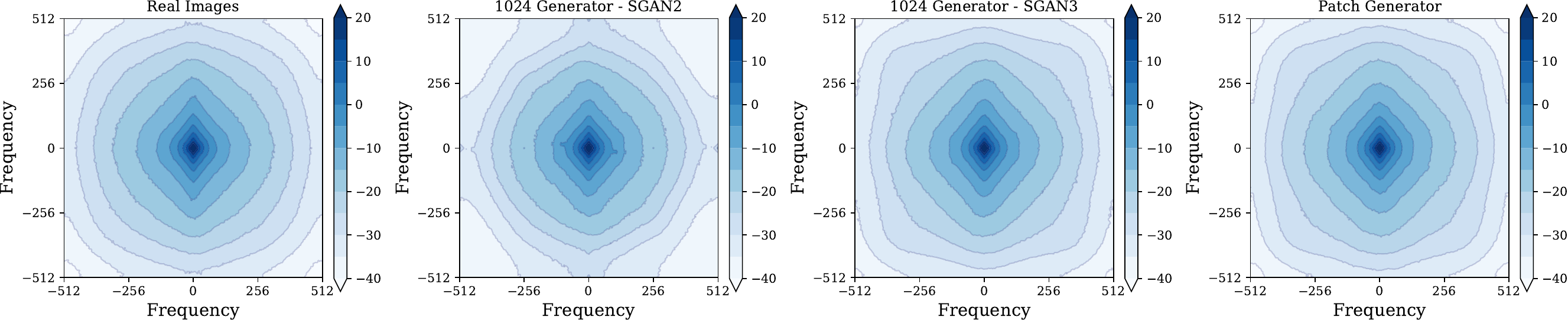}
\includegraphics[width=\textwidth]{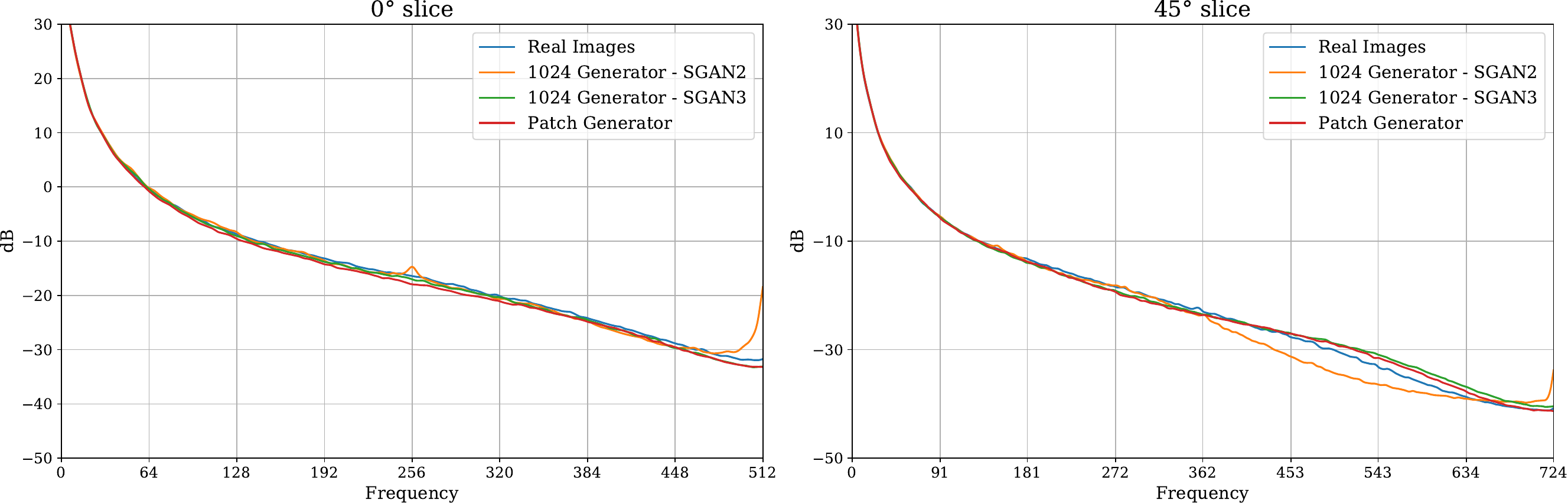}
\caption{\label{fig:sm_fft}\small Comparison of frequency distribution. We plot the frequency spectrum of real images, StyleGAN2 and StyleGAN3 trained on the entire FFHQ dataset at 1024 resolution, and our Patch Generator which is trained on $\PatchResolution \times \PatchResolution$ patches of FFHQ6K (which contains approximately 1k images at 1024 resolution and 5k at lower resolutions). The frequency distributions are similar, suggesting that even a smaller generator is able to approximate fine textures well.}
\end{figure}

Despite being trained to generate patches, our generator can approximately match
the frequency content in real images, and that of a StyleGAN3 model trained for
1024 resolution generation on the full FFHQ dataset (Fig.~\ref{fig:sm_fft}).
While StyleGAN2 achieves better FID than StyleGAN3, we find that it has a
different frequency profile that is less similar to that of real images. We
compare the FID of these oracle models with our continuous patch model in
Tab.~\ref{tab:sm_oracle}. While the oracles can achieve lower FID and pFID variants, we note that
training the oracle assumes that a sufficient number of high-resolution images
of the same size are available, and trains the model specifically for a fixed
resolution, whereas we employ mixed resolution training on fewer than 10\% of
the full HR dataset. Our training strategy therefore allows us to take advantage of the
varied resolutions of images in the wild, which is not possible in the oracle setting. 

\subsection{Additional Model Variations}\label{sec:sm_model_variations}

In Section~\ref{sec:expt_model_variations} of the main text, we describe and study variations of our model. Here, we provide additional quantitative and qualitative results and study additional factors.

\begin{table}[t]
\parbox{0.48\linewidth}{
\caption{ \small We evaluate additional precision and recall metrics~\cite{sajjadi2018assessing}, and their corresponding patch variants, for our model, naive upsampling, and super-resolution methods on the FFHQ dataset. We also include pFID at a fixed 1024 scale evaluated in Tab.~\ref{tab:sm_oracle}. \label{tab:sm_ffhq_addtl_metrics}}
\centering
  \resizebox{0.95\linewidth}{!}{
  \begin{tabular}{lccccc}
    \toprule
    & \multicolumn{2}{c}{Precision} & \multicolumn{2}{c}{Recall} & \multicolumn{1}{c}{pFID} \\
    \cmidrule(lr){2-3}  \cmidrule(lr){4-5}  \cmidrule(lr){6-6}
    & 1024 & Patch & 1024 & Patch & 1024 \\
    \midrule
    Upsample & 0.68 & 0.77 & 0.37 & 0.20 & 31.70 \\
    Real-ESRGAN & 0.40 & 0.54 & 0.51 & 0.47 & 20.04 \\
    GFP-GAN & 0.48 & 0.62 & 0.34 & 0.32 & 20.58 \\
    Ours & 0.69 & 0.68 & 0.47 & 0.55 & 8.01 \\
    \bottomrule
  \end{tabular}
  }
}
\hfill
\parbox{.48\linewidth}{
\caption{ \small Variations of teacher regularizer on FFHQ6k. The inverse teacher regularization outperforms forward regularization, and adding a scale-conditioning branch further improves  higher resolutions. Omitting the teacher harms global structure. (*) indicates  default setting. \label{tab:teacher}}
\centering
  \resizebox{0.9\linewidth}{!}{
  \begin{tabular}{lcccc}
    \toprule
    & \multicolumn{3}{c}{FID} &pFID \\
    \cmidrule(lr){2-4}\cmidrule(lr){5-5} 
    & 256 & 512 & 1024 & random \\

    \midrule
    Forward teacher & 3.23 & 4.21 & 5.35 & 6.06 \\
    Inverse teacher  & 3.35 & 4.18 & 4.88 & 4.67 \\
    No teacher      & 5.50 & 5.93 & 7.13 & 3.17 \\
    Inverse + scale (*) & 3.37 & 4.41 & 4.47 & 4.28 \\
    \bottomrule
  \end{tabular}
  }
}
\end{table}

\begin{figure}[!tbh]
\centering
\includegraphics[width=\textwidth]{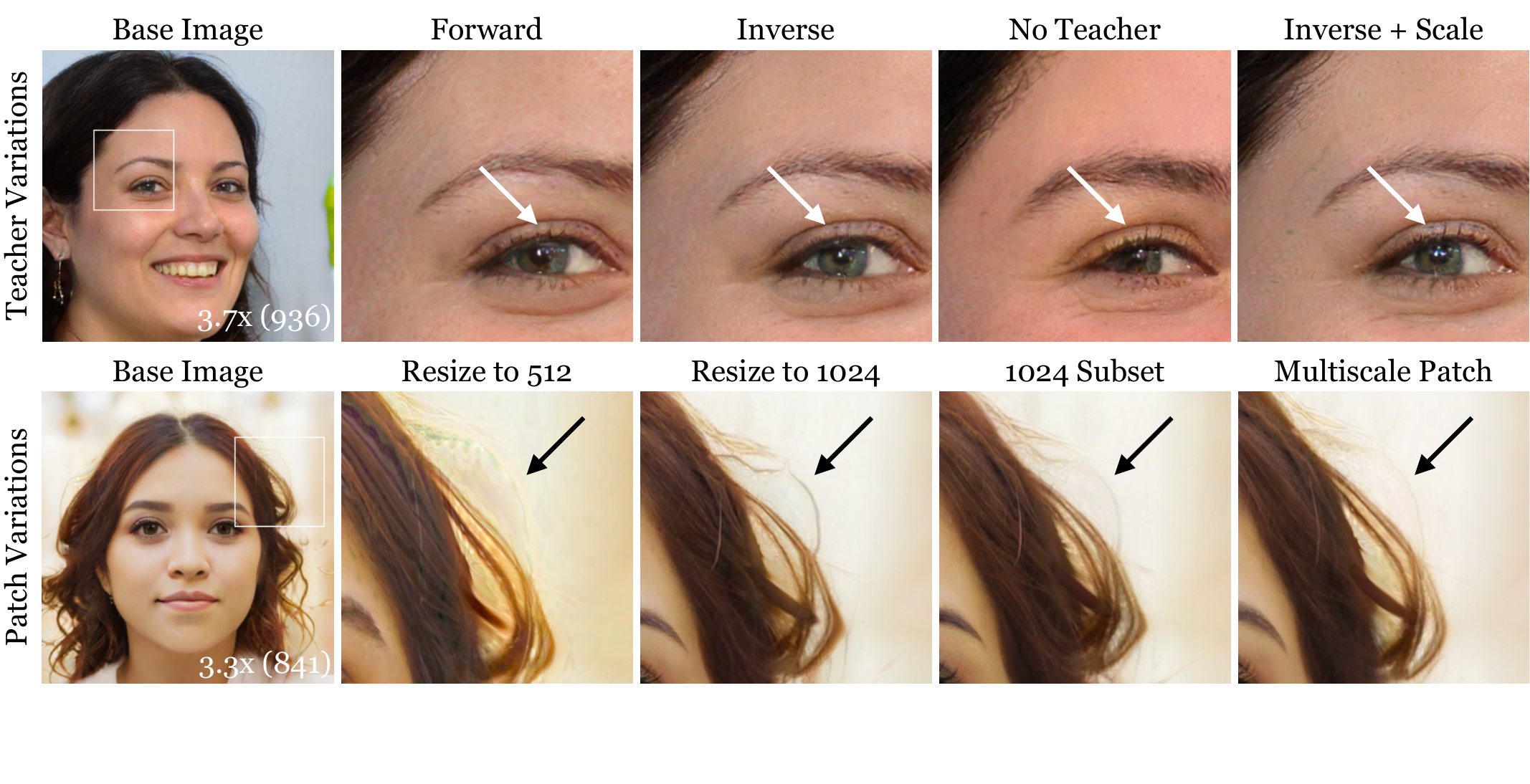}
\vspace{-10mm}
\caption{\small Qualitative examples of model variations. (Top) Using an inverse teacher loss and the scale conditioning branch generates sharper details while also preserving similarity to the base image. (Bottom) We compare our multi-size training approach to methods that
do not take advantage of different image sizes and instead train for a fixed resolution. Fixed-resolution training cannot generalize to other resolutions, and upsampling images leads to blurring. Our final model is able learn from mixed-resolution training images and also synthesize at arbitrary resolutions.}
\label{fig:sm_variations}
\end{figure}

\begin{table}[t]
\parbox{.53\linewidth}{
\caption{ \small Teacher regularization weight trades off between improved detail synthesis (pFID) and global realism (full image FIDs).
We choose an in-between value ($\lambda_\text{teacher}=5$); this value can be adjusted based on desired similarity to the base resolution. (*) indicates our default setting. \label{tab:teacher_weight}
}
\centering
    \resizebox{0.8\linewidth}{!}{
  \begin{tabular}{lccccc}
    \toprule
    & \multicolumn{3}{c}{FID} & pFID  & \multirow{2}{*}{L1}\\
    \cmidrule(lr){2-4}\cmidrule(lr){5-5} 
    & 256 & 512 & 1024 & random \\

    \midrule
    $\lambda_\text{teacher}=0$        & 5.50 & 5.93 & 7.13 & 3.17 & 0.16\\
    $\lambda_\text{teacher}=2$        & 3.42 & 4.58 & 5.46 & 3.15 & 0.10\\
    $\lambda_\text{teacher}=5$ (*)      & 3.37 & 4.41 & 4.47 & 4.28 & 0.08 \\
    $\lambda_\text{teacher}=10$       & 3.46 & 4.25 & 4.61 & 5.39 & 0.07\\
    \bottomrule
  \end{tabular}
  }
}
\hfill
\parbox{.45\linewidth}{
\caption{ \small We sample patches from the HR dataset at global resolutions between $
(s_\text{min}, s_\text{max})$. The same model architecture trained on patches from higher resolution images improves the synthesis result at 1024 resolution. (*) indicates our default setting.\label{tab:sm_extrapolation}
}
\centering
    \resizebox{1.0\linewidth}{!}{
  \begin{tabular}{lcccc}
    \toprule
    & \multicolumn{3}{c}{FID} &pFID \\
    \cmidrule(lr){2-4}\cmidrule(lr){5-5} 
    & 256 & 512 & 1024 & random \\

    \midrule
    (256, 512)      & 5.20 & 5.92 & 19.01 & 35.66 \\
    (256, 1024)     & 3.43 & 4.16 & 4.61 & 4.19 \\
    (512, 1024) (*)    & 3.28 & 4.04 & 4.16 & 3.61 \\
    \bottomrule
  \end{tabular}
  }
}
\end{table}
\myparagraph{Alternative metrics}
In addition to FID and pFID, we also report precision and recall~\cite{sajjadi2018assessing} for our model, naive upsampling, and super-resolution models (Tab.~\ref{tab:sm_ffhq_addtl_metrics}). Super-resolution obtains lower precision (suggesting out-of-distribution results) and similar or lower recall. Upsampling obtains similar or better precision, but lower recall (suggesting that is does not sufficiently cover the real image distribution). Here, we also include pFID measured at the maximum resolution for FFHQ (1024).

\myparagraph{Variations on teacher regularization.} In the main text, we
introduce variations on the teacher regularization including ``forward'' and
``inverse'' loss formulations, and discarding the teacher regularization
all-together. Tab.~\ref{tab:teacher} shows the FID comparisons of these three
variants, in which the ``inverse'' loss obtains the best FID scores at the
highest 1024 resolution. Adding the scale-conditioning branch to inject scale
information throughout the generator further improves FID@1024 and pFID. We show
qualitative examples in Fig.~\ref{fig:sm_variations} (top), where the inverse teacher
with scale-conditioning input can synthesize the cleanest details while still
being similar to the base image.

As default, we set $\lambda_\text{teacher}=5$ during the patch-based training
phase. Changing $\lambda_\text{teacher}$ balances between local image quality
and similarity to the base resolution image, where higher
$\lambda_\text{teacher}$ offers the most similarity to the base resolution with
lower L1 difference, but lower $\lambda_\text{teacher}$ improves pFID,
suggesting better quality of the synthesized patches
(Tab.~\ref{tab:teacher_weight}). 

\begin{figure}[!tbh]
\centering
\includegraphics[width=\textwidth]{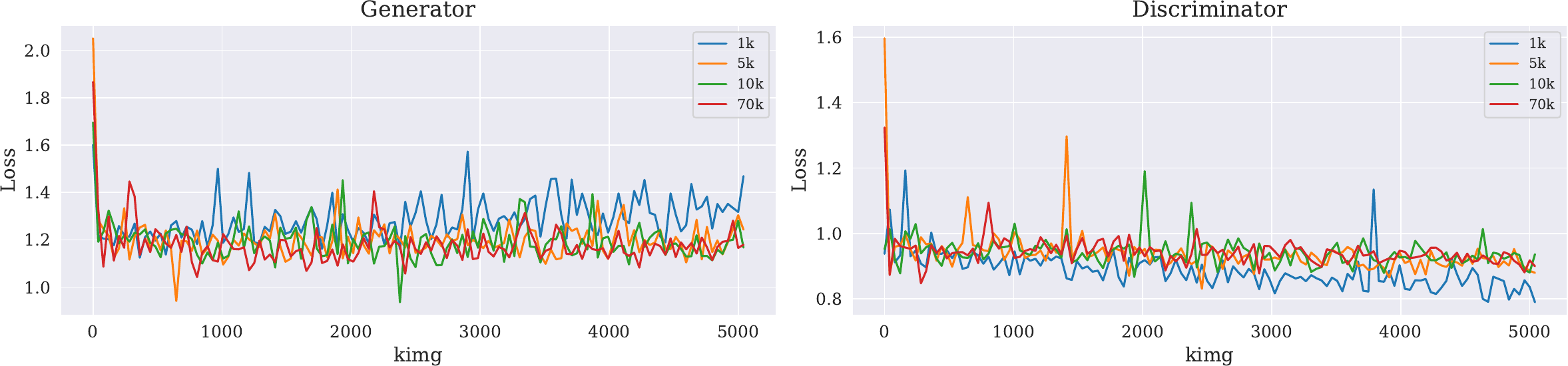}
\caption{\small Number of training images. While FID numbers are similar, we find that using 1K HR images for training shows some evidence of divergence. The training dynamics of 5K images is similar to that of the full dataset. }
\label{fig:dataset_size}
\end{figure}

\myparagraph{Fixed-size vs Multi-size training.} Fig~\ref{fig:sm_variations} (bottom)
shows an example of a synthesized patch comparing our multi-size training to
strategies of fixed-size training. Fixed-size training does not naturally
generalize to other sizes, causing deterioration in image quality when sampled
at resolutions not equal to the training resolution. Upsampling the training
images to a common resolution introduces blurriness in the synthesized output.
The result of training on only the subset of images at 1024 resolution looks
qualitatively similar to that of multi-scale training, but multi-scale training
attains better FID metrics and is able to use more images for training.

\myparagraph{Changing the number of training images.} While the model FID scores
remain largely similar (within a range of 0.3) when training on 1k to 70k
high-resolution images, we found that using 1k images showed some evidence of
training divergence (Fig.~\ref{fig:dataset_size}). On the other hand, the
training trajectory of using 5k images looks largely similar to that of using
the full HR dataset (70k) images. Therefore, when collecting images for the
remaining domains, we aim to collect between 5k-10k images to construct the HR
dataset. 

\begin{figure}[ht!]
\centering
\includegraphics[width=\textwidth]{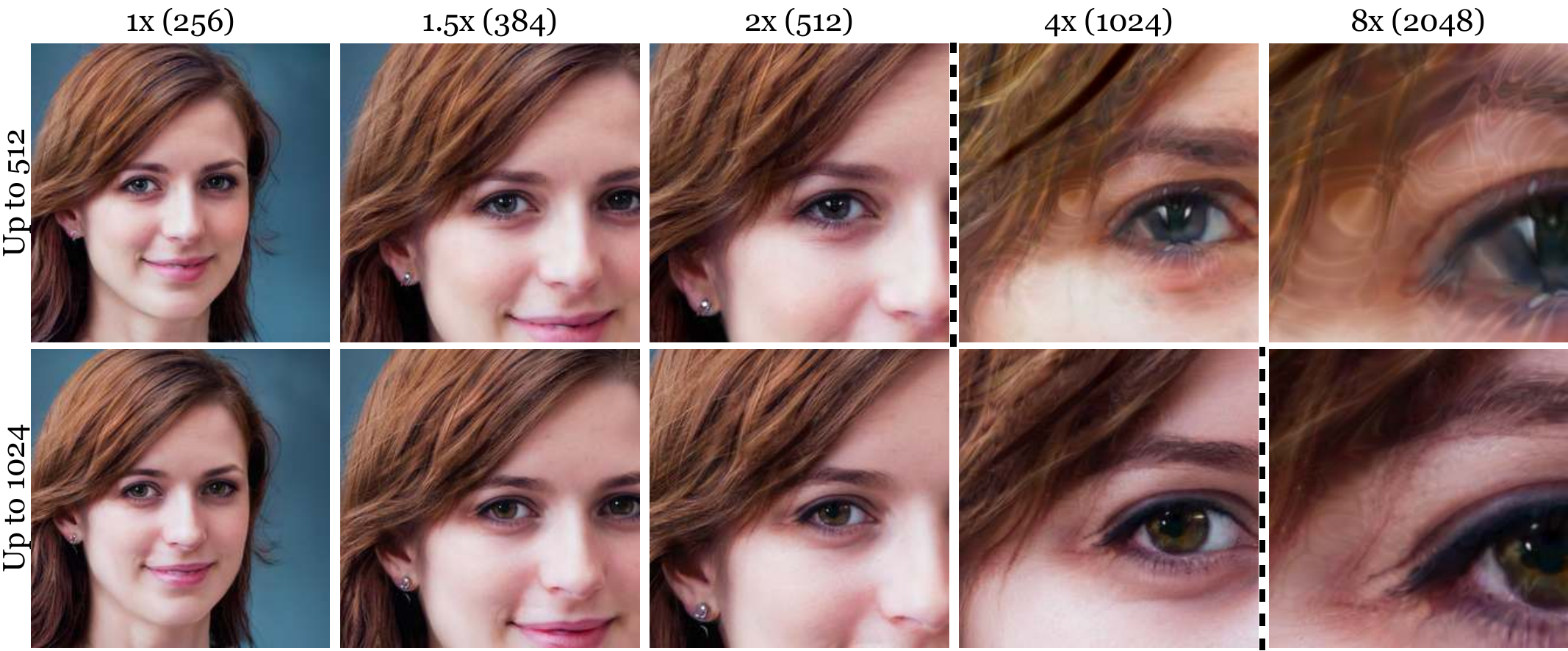}
\caption{\label{fig:extrapolation}\small Impact of sampling resolutions. Using the same model architecture and FFHQ6k dataset, we sample images (top) from 256 to 512 resolution, and (bottom) from 512 to 1024 resolution. The dotted line indicates when inference resolution exceeds the maximum training resolution. Watery artifacts start to appear when extrapolation, but this can be tempered by simply training on patches from larger images.}
\end{figure}

\myparagraph{Investigating the impact of sampling resolutions.} Our FFHQ6k
dataset contains images between 512 and 1024 resolution, and during training the
images are randomly downsampled from their native resolution, and can be
optionally clipped at an upper resolution. Here, we conduct experiments to study
the effects of these sampling ranges. When training the model on resolutions
\FullImageResolution sampled between 256 and 512, the image quality declines by
1024 resolution at inference time and contains visual artifacts
(Tab.~\ref{tab:sm_extrapolation}, Fig.~\ref{fig:extrapolation}). Taking the same
image and model architecture, but instead training on resolutions between 256 and 1024
offers better FID@1024, and sampling from 512 to 1024 resolution further
improves FID@1024. As before, all models are trained on patches of size
$\PatchResolution=256$, and the model is jointly trained on the fixed-size
dataset to preserve FID@256. Accordingly for the other domains, our sampled
resolutions for HR dataset range between the native resolution
$\NativeResolution$ and the minimum resolution of the HR images. These results
suggest that the synthesized resolution can be dictated by the training images;
simply adding patches from higher resolution images can allow the same model to
better synthesize at a higher resolution.
We also tried an experiment using a separately trained smaller 128px patch generator and the same 512 to 1024 resolution patch sampling scheme, but obtained worse FID (8.38 at 1024 resolution compared to 4.16 for our default model); we hypothesize this is because may be due to worse FID from the initial pretraining phase that carries over to the patch training phase (5.14 compared to 3.71 for our default model).

\begin{wraptable}{r}{6.5cm}
\vspace{-8mm}
\parbox{1.0\linewidth}{
\caption{ \small Discriminator variations. Our default discriminator, which jointly trains globally on the LR dataset and patches from the HR dataset attains the best FID metrics. Other changes to the discriminator did not improve performance. (*) indicates our default setting. \label{tab:sm_disc_variations}}
\centering
  \resizebox{1.0\linewidth}{!}{
  \begin{tabular}{lcccc}
    \toprule
    & \multicolumn{3}{c}{FID} &pFID \\
    \cmidrule(lr){2-4}\cmidrule(lr){5-5} 
    & 256 & 512 & 1024 & random \\

    \midrule
Default Discriminator (*) & 3.28 & 4.04 & 4.16 & 3.61 \\
No Base Resolution & 9.96 & 4.23 & 4.69 & 2.92 \\
Two Discriminators & 3.82 & 4.63 & 5.34 & 3.38 \\
Scale-conditioned Discriminator & 31.81 & 71.33 & 89.38 & 120.06 \\
    \bottomrule
  \end{tabular}
  }
}
\vspace{-6mm}
\end{wraptable}

\myparagraph{Changing the discriminator.} Our final model introduces changes to the generator, but keeps the same discriminator from the initial pretraining step. During patch-training, the discriminator must also learn to distinguish between real and synthesized patches. Here, we investigate alternatives of changing the discriminator setup (Tab.~\ref{tab:sm_disc_variations}). (1) We remove sampling from the LR dataset, now causing the discriminator to focus entirely on patches. This causes pFID to improve but the remaining global FIDs to worsen. In particular, this allows the generator to forget how to synthesize at the base resolution, causing a large increase in FID@256. The impact of sampling from the base resolution and the teacher regularization have similar outcomes: both encourage global coherence, but the teacher has a stronger effect than base resolution training. (2) We also try adding a second discriminator so that one focuses entirely on the global low-resolution image, and the other entirely on patches. Both discriminators are initialized with the result from pretraining, but we find that this setting leads to suboptimal metrics, compared to using a single discriminator. (3) We inject scale information into the discriminator following a similar method as the generator via weight modulation. In this case, the training becomes unstable as the discriminator is able to out-compete the generator. 

\myparagraph{Removing the pretraining step.} %
Our final model is first pretrained at a fixed, smaller resolution before varied-size patch training is enabled. This pretraining phase encourages global coherence and also serves as the teacher model later during patch-based training. We conduct an experiment in which the initial pretraining step is omitted, and the model is trained on randomly sampled patches from the start of training. When trained with the same number of HR image patches, the model without global pretraining suffers in both structure and texture -- FID for 1024px generated images is 26.78 and pFID is 8.64 -- compared to our original model which performs global pretraining at low resolution -- FID at 1024px is 4.50 and pFID is 3.46 after 10M training images.

\subsection{Detectability}

\begin{wrapfigure}{r}{0.5\textwidth}
    \vspace{-12mm}
  \begin{center}
        \includegraphics[width=0.45\textwidth]{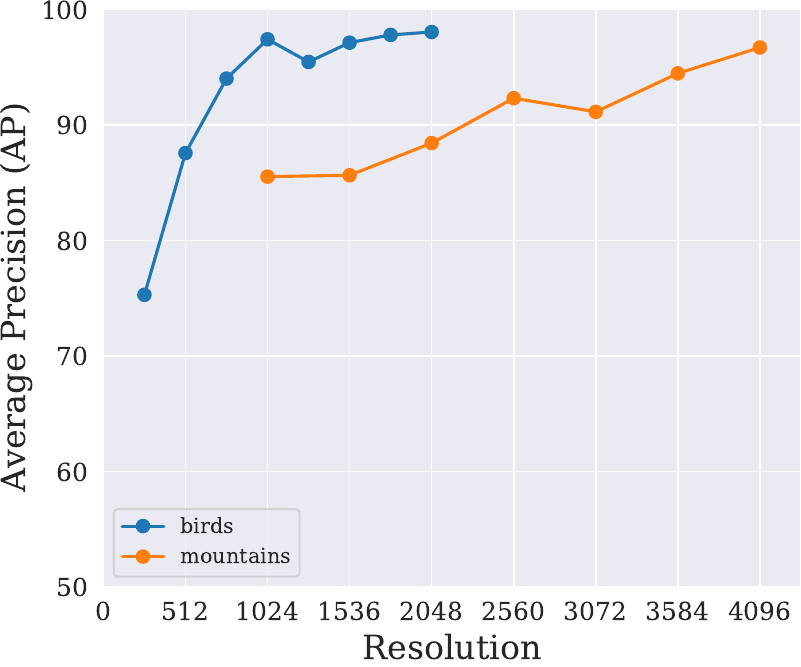}
  \end{center}
  \vspace{-6mm}
    \caption{\small Detection score from~\cite{wang2020cnn} on our Birds and Mountains datasets. All scores are above chance of $50\%$. Note in both cases, the detectability of our network trends upwards with resolution.\label{fig:detectability}}
  \vspace{-10mm}
\end{wrapfigure}

A concern with improved image generation is the potential for more convincing deceiving images, particularly those of higher resolution, which is the focus of our work.
We use the off-the-shelf detector from Wang et al.~\cite{wang2020cnn} on our Birds (generated $256\rightarrow 2048$) and Mountains ($1024\rightarrow 4096$) generators, across a large range of resolutions. As shown in Figure~\ref{fig:detectability}, the scores are well above chance ($50\%$) across both datasets and resolutions. Interestingly, the curve generally trends upwards, indicating that while higher resolution images may look more natural, they are also easier to detect.

\section{Additional implementation details}\label{sec:sm_method}

Building off the StyleGAN3~\cite{karras2021alias} architecture, we describe our coordinate conditioning and scale modulation branch applied to enable generation of multi-scale patches during the second training phase.

\subsection{Patch-based training}

\myparagraph{Extracting patches from varied size images} From our dataset of images \Dataset, we sample
an image $x_i\in \mathds{R}^{H_i\times W_i\times 3} \sim
\Dataset$, with short-side $\NativeResolution=\min(H_i,W_i)$, and take an \NativeResolution-by-\NativeResolution
square crop. We then Lanczos downsample the image to an intermediate resolution $\FullImageResolution \in [\PatchResolution, \NativeResolution]$, which provides ``free'' additional views from the same image, without
introducing image corruptions. 

Next, we sample a random crop of size
\PatchResolution and record the sampling location $v\in \mathds{R}^2$.
To summarize this procedure, we obtain a patch $\ImageSample\in
\mathds{R}^{\PatchResolution \times \PatchResolution \times 3}$ from these two
operations, while saving the sampled image resolution \FullImageResolution and patch center location
$\PatchLocation=(\PatchLocationX, \PatchLocationY)\in [0, 1]^2$ for later use.
\begin{equation}
\ImageSample, \FullImageResolution, \PatchLocation = \mathrm{Crop} ( \mathrm{Downsample} (x_{i}))
\end{equation}

\myparagraph{Synthesizing patches from the generator} Given a set of patches from the image dataset, the generator is tasked with synthesizing images at the corresponding patch locations. To transform the normalized coordinate domain \ContinuousDomain into patch coordinates, we apply a transformation matrix to each 2D location \NormalizedCoordinates in homogenous coordinates:
\begin{equation}
 \Coordinates =  \TransformPatch * \NormalizedCoordinates = 
 \begin{bmatrix}
\frac{\PatchResolution}{\FullImageResolution} & 0 & \PatchLocationX \\
0 & \frac{\PatchResolution}{\FullImageResolution} & \PatchLocationY \\
0 & 0 & 1 \\
\end{bmatrix}
* \NormalizedCoordinates
\end{equation}

Following StyleGAN3, these transformed coordinates are then encoded as
$K$ random Fourier channels by multiplying by frequencies $B\in \mathds{R}^{K
\times 2}$ and adding phases $\phi\in \mathds{R}^{K}$.
For patch synthesis, the Fourier feature extraction at index $(h,w)$ becomes:
\begin{equation}
F_{h,w}(\Coordinates)=\sin \big(2\pi \hspace{.5mm} B \Coordinates + \phi \big) \in \mathds{R}^{K},
\end{equation}

\subsection{Scale-conditioning branch}

As individual coordinate positions \Coordinates do not directly convey scale information, we found it beneficial additionally incorporate the scale input to intermediate layers of the generator. To do this, we first normalize the target scale \FullImageResolution to $[0, 1]$ using:
\begin{equation}
    \NormalizedFullImageResolution = \frac{\FullImageResolution - \PatchResolution}{\MaxNormImageResolution - \PatchResolution} 
\end{equation}
where \MaxNormImageResolution is selected from dataset statistics and is only present as a normalization factor, but does not clip the upper synthesis bound during inference. 
Empirically, we found that adding a small offset factor (we use $0.1$) to the normalized target scale \NormalizedFullImageResolution allows for smoother interpolations between resolutions by avoiding a discontinuity at zero.

We then encode \NormalizedFullImageResolution using a parallel mapping network of identical architecture to the latent mapping network $\AuxNetwork(\LatentCode)$, and add the two inputs after undergoing a layer-specific affine transformation into style-space~\cite{wu2021stylespace,patashnik2021styleclip} to obtain the final modulation parameter $\AuxNetwork(\LatentCode, \FullImageResolution)_k$ at layer $k$:
\begin{equation}
   \AuxNetwork(\LatentCode, \FullImageResolution)_k = (W_{z, k} * \AuxNetwork_z(\LatentCode) + b_{z, k}) +  (W_{s, k} * \AuxNetwork_s(\FullImageResolution) + b_{s, k})
\end{equation}
Because the modulation parameter is a multiplicative factor on the network weights and the scale-conditioning portion is added only during the secondary patch-wise training step, we initialize $b_{z, k} = \mathbf{1}$ and $b_{s, k} = \mathbf{0}$ to allow the network to smoothly transition between the initial pretraining step and secondary patch-based training.

\subsection{Training procedure}

We train our models on four to eight V100 GPUs
with 16GB memory.
By sampling fixed-size patches, the memory and compute
footprint remain constant during training.
For FFHQ, we finetune our initial fixed-scale generator from the pretrained
FFHQ-U model~\cite{karras2021alias}, which reaches a minimum FID within 4M
training images.
In the remaining domains, we perform the pretraining step
from scratch, retaining the checkpoint with the lowest
FID, computed over 25M image samples, before
continuing with the second, mixed-resolution training phase.
Our training procedure is compatible with both 3$\times$3 and 1$\times$1 kernel
sizes in StyleGAN3 (T \& R configurations, respectively). For the patch-based training step, we proceed with the model configuration that reaches the best FID in pretraining, which is typically Config T with the exception of the FFHQ domain.

\end{document}